\documentclass[11pt]{article}
\usepackage[margin=1in]{geometry}
\usepackage{graphicx}
\usepackage{authblk}
\usepackage{booktabs}
\usepackage{longtable}
\usepackage{hyperref}
\usepackage{array}
\usepackage{url}
\usepackage{amsmath}
\usepackage{caption}
\usepackage{multirow}
\usepackage{url}
\usepackage{cite}
\usepackage{float}
\usepackage{tabularx}
\usepackage[inkscapelatex=false]{svg}

\graphicspath{{figures/}}

\title{Large Language Model-Based Uncertainty-Adjusted Label Extraction for Artificial Intelligence Model Development in Upper Extremity Radiography}
\author[1,2]{Hanna Kreutzer \thanks{Corresponding author: \href{mailto:hannkreutzer@ukaachen.de}{hannkreutzer@ukaachen.de}}}
\author[1,2]{Anne-Sophie Caselitz}
\author[3]{Thomas Dratsch}
\author[4]{Daniel Pinto dos Santos}
\author[2]{Christiane Kuhl }
\author[1,2]{Daniel Truhn }
\author[1,2]{Sven Nebelung }

\affil[1]{Lab for Artificial Intelligence in Medicine, Department of Diagnostic and Interventional Radiology, University Hospital Aachen, Aachen, Germany}
\affil[2]{Department of Diagnostic and Interventional Radiology, University Hospital Aachen, Aachen, Germany}
\affil[3]{Institute for Diagnostic and Interventional Radiology, Faculty of Medicine and University Hospital Cologne, University of Cologne, Cologne, Germany}
\affil[4]{Department of Diagnostic and Interventional Radiology, University Medical Center Mainz, Mainz, Germany}

\date{}

\begin{document}
\maketitle

\begin{abstract}
\textbf{Objectives:} To evaluate GPT-4o’s zero-shot ability to extract structured diagnostic labels (with uncertainty) from free-text radiology reports and to test how these labels affect multi-label image classification of musculoskeletal radiographs.

\textbf{Materials and Methods:} This retrospective study included radiography series of the clavicle (n=1{,}170), elbow (n=3{,}755), and thumb (n=1{,}978). After anonymization, GPT-4o filled out structured templates by indicating imaging findings as present (``true''), absent (``false''), or ``uncertain.'' To assess the impact of label uncertainty, ``uncertain'' labels of the training and validation sets were automatically reassigned to ``true'' (inclusive) or ``false'' (exclusive). Label--image pairs were used for multi-label classification using a ResNet50. Label extraction accuracy was manually verified on internal (clavicle: n=233, elbow: n=745, thumb: n=393) and external test sets (n=300 for each). Performance was assessed using macro-averaged receiver operating characteristic (ROC) area under the curve (AUC), precision recall curves, sensitivity, specificity, and accuracy. AUCs were compared with the DeLong test.

\textbf{Results:} Automatic extraction was correct in 98.6\% (60{,}618 of 61{,}488) of labels in the test sets. Across regions, model training yielded competitive macro-averaged AUC values for inclusive (e.g., elbow: AUC=0.80 [range, 0.62--0.87]) and exclusive models (elbow: AUC=0.80 [range, 0.61--0.88]). Models generalized well on external datasets (elbow [inclusive]: AUC=0.79 [0.61--0.87]; elbow [exclusive]: AUC=0.79 [0.63--0.89]). No significant differences were observed across labeling strategies or datasets ($p\geq0.15$).

\textbf{Conclusion:} GPT-4o extracted labels from radiologic reports to train competitive multi-label classification models with high accuracy. Detected uncertainty in the radiologic reports did not influence the performance of these models. 

\end{abstract}

\section{Introduction}
An estimated 3.6 billion imaging procedures are conducted globally each year, according to the World Health Organization \cite{WorldHealthOrganization.14.April2016}. Despite this abundance, AI model development is challenged by data scarcity \cite{Piffer.2024}. Increasing the availability of imaging study-report pairs would enhance data quantity and variability for AI model training and help to address performance issues, such as weak generalizability \cite{Yu.2022}. Traditionally, such datasets are created through manual annotation, an approach that is labor-intensive, costly, and, if performed by non-experts, inconsistent in quality \cite{Sylolypavan.2023}. Automated label extraction from radiologic reports has also been attempted using rule-based or conventional machine-learning natural language processing (NLP) methods \cite{CPereira.2024}. While promising \cite{PintoDosSantos.2019,Stember.2022,Irvin.2019,Dai.2024}, these require additional training and often struggle with terminological complexity Mislabeling rates of up to 10\% have been reported \cite{Sun.2022}, potentially degrading model performance.\\
Large Language Models (LLMs) offer a more adaptable alternative for processing radiologic reports, with the ability to interpret complex linguistic patterns and transform free text into structured templates \cite{Adams.2023,Truhn.2024}. For instance, Al Mohamad et al. used an open-source LLM (Mixtral 8×7b) to extract binary fracture labels from ankle radiographs and train a standard Convolutional Neural Network (CNN) \cite{AlMohamad.2025}. However, their study addressed a single label in one anatomic region and did not account for label uncertainty, i.e., phrases such as “likely,” “suggestive,” or “indeterminate”, that are common in radiology reports \cite{Callen.2020} and reflect inherent diagnostic ambiguity. Ignoring uncertainty risks introducing noise into datasets, potentially impairing downstream AI performance.\\
To our knowledge, no prior study has explored automated, uncertainty-aware label extraction using LLMs across multiple upper-extremity regions for training multi-label classification models. While fracture-detection AI models in the upper extremity have been reviewed \cite{Kuo.2022}, comprehensive models addressing both common and less frequent conditions of the clavicle, elbow, and thumb remain understudied.\\
Our objective was to investigate whether LLMs can be used for automated label extraction across multiple anatomic regions of the upper extremity to train multi-label classification models and whether label uncertainty impacts model performance. We hypothesized that (i) LLMs can accurately extract labels from radiologic reports while detecting uncertainty, (ii) label uncertainty does not affect model performance, and (iii) extracted labels enable efficient training of multi-label classification models.

\section{Methods}
\subsection{Study Design}
This study was designed as a two-center retrospective analysis utilizing radiography series and original radiologic reports of the elbow, thumb, and clavicle. The internal dataset from our university hospital (Department of Diagnostic and Interventional Radiology, University Hospital Aachen) was sourced from our local Picture Archiving and Communication System (isite, Philips) spanning 2010 to 2024. The study was conducted in accordance with local data protection laws, following approval from the local ethical committee (Ethical Committee, Medical Faculty, RWTH Aachen University, EK24-174), with a waiver for individual informed consent. External test sets were collected from the University Hospital Cologne (Institute of Diagnostic and Interventional Radiology) with ethical approval (24-1348-retro) and a waiver for informed consent. Radiographs at the internal site were acquired using five computed-radiography units: Philips DigitalDiagnost VR, Philips DigitalDiagnost 4 High Performance, Siemens Ysio Max, and two Philips DigitalDiagnost C90 systems. Radiographs at the external site were acquired using a Philips DigitalDiagnost system. All units employed flat-panel detectors, automatic exposure control, and the site’s standard radiography protocols. Routine upper-extremity exposure settings were 50–60 kVp and 2–5 mAs for the elbow and thumb, and 60–70 kVp, 8–18 mAs for the clavicle, using a 100 cm source-to-image distance.

\subsection{Patient Selection}
\begin{figure}[H]
  \centering
  \includegraphics[width=0.8\linewidth]{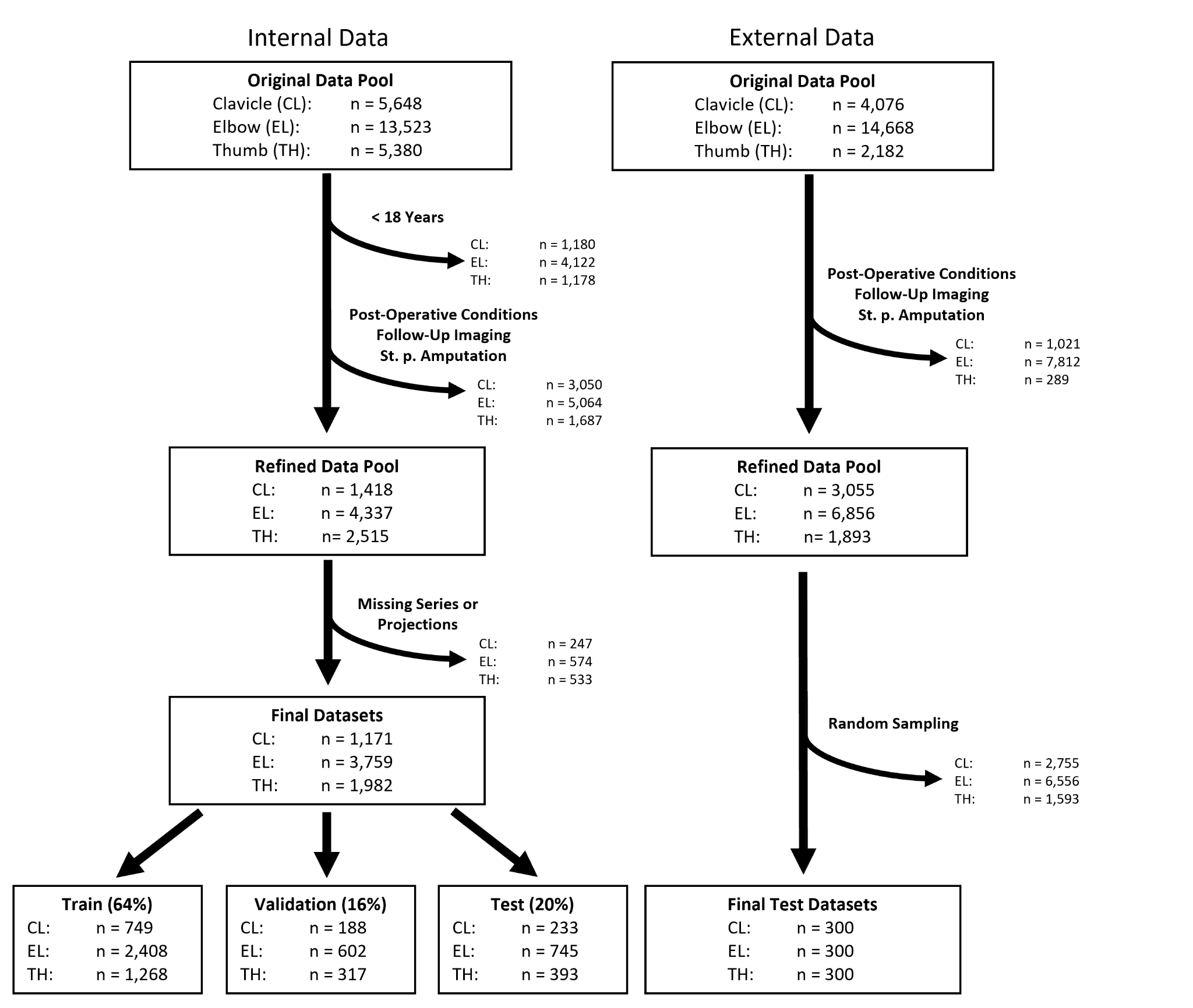}
  \caption{Data Curation and Preparation. 
Left: internal dataset (University Hospital Aachen, 2010–2024); right: external dataset (University Hospital Cologne, 2010-2022). Identical exclusion criteria were applied to both sources: patients $<$18 years, post-operative imaging, follow-up examinations, and studies after amputation. Pediatric cases had already been removed by the Cologne site (note beneath the age-exclusion box). After exclusions, the Cologne pool underwent region-stratified random sampling to 300 studies each of the clavicle (CL), elbow (EL), and thumb (TH). The internal data were split into training (64\%), validation (16\%), and internal test (20\%) subsets. The external data served only for final testing. All final datasets comprise anteroposterior projections of the clavicle and both anteroposterior and lateral projections of the elbow and thumb.
}
  \label{fig:flow}
\end{figure}

Patient selection is visualized in Figure~\ref{fig:flow}. The final internal datasets were split into training (64\%), validation (16\%), and test sets (20\%) while ensuring balanced label distribution across the sets. Because this retrospective study used all examinations that met the inclusion criteria, no a-priori sample-size calculation was performed. For the external data, the collaborating hospital had already removed studies of patients \textless 18 years and examinations with missing projections. Therefore, only post-operative, follow-up, and amputation cases had to be excluded locally. An automated Python script (v 3.12.3) then drew a random sample of 300 patients per anatomic region for the final external test sets

\subsection{Overall Approach}
\begin{figure}[H]
  \centering
  \includegraphics[width=0.6\linewidth]{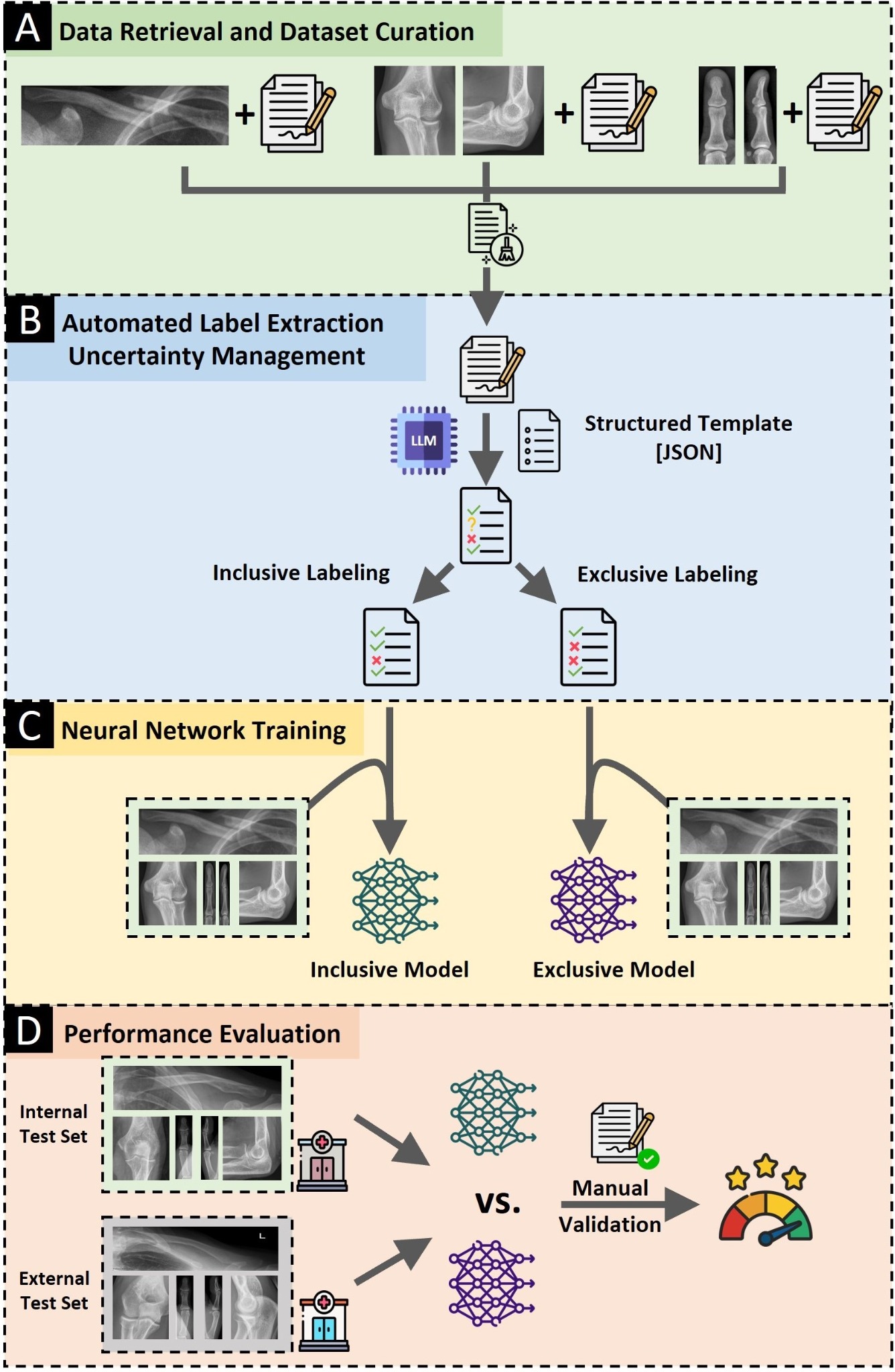}
  \caption{Study Workflow.
A) Radiography series of the clavicle, elbow, and thumb and corresponding radiologic reports were collected and curated. B) The LLM filled out a region-specific structured template containing relevant conditions based on the radiologic reports. Individual labels were either “true,” false,” or “uncertain.” The template was machine-readable and available in the JavaScript Object Notation (JSON) format. For subsequent model training, “uncertain” labels of the training/validation sets were automatically  converted into “true” (inclusive labeling) or “false” (exclusive labeling) using Python. C) The JSON files were paired with the radiography series to train the image classification models as inclusive and exclusive versions using the respective labels. D) Both models were tested on internal and external test datasets containing manually corrected labels (ground truth).
}
  \label{fig:overview}
\end{figure}

The general procedure is illustrated in Figure~\ref{fig:overview}. OpenAI’s GPT-4o was used to extract structured labels from free-text radiologic reports, which were used to train dedicated CNNs for classification. Python was used for data processing and implementation of the models.

\subsection{Automatic Label Extraction}
Before label extraction, reports were automatically anonymized by removing personal identifiers such as patient names, physician details, dates, and patient IDs. Successful anonymization was manually verified. GPT-4o was used in a zero-shot manner, i.e., operated without prior task-specific fine-tuning on label extraction (Supplementary Text ~\ref{suppl_text_1}). The tool was instructed to fill out predefined JavaScript Object Notation (JSON) templates for each anatomic region that contained common conditions like fractures and less common findings like sclerotic lesions and soft tissue masses (Supplementary Text ~\ref{suppl_text_2}). The templates were designed by a senior MSK radiologist (S.N., with 12 years of clinical experience).
For each label (pathology), the tool was instructed to choose from three options: “true”, “false”, or “uncertain.” “True” indicates that the imaging finding of interest is present (pathology present), “false” indicates that the finding is absent (pathology absent), and “uncertain” indicates that the report contains expressions of diagnostic uncertainty. Consequently, labels were marked as “uncertain” when reports contained phrases indicating doubt or hedging terms. The uncertainty terms were based on dedicated literature \cite{Callen.2020,Panicek.2016} and included expressions such as “possibly” and “suspected.” These examples were provided along with their German translations since the reports were in German. 

\subsection{Manual Quality Check}
Following the automated label extraction, a medical student (A.C.) was trained and supervised by the senior MSK radiologist (S.N.) to manually verify and correct the labels for the internal and external test sets. The verification process involved comparing the extracted labels with the original reports. “Uncertain” labels were not allowed in the test, and thus, follow-up imaging studies were reviewed to definitively categorize each label as “true” (pathology present) or “false” (pathology absent).

\subsection{Model Training}
Model training was implemented using PyTorch (version 2.4.0) \cite{Paszke.2019}. For each anatomic region, we trained two model variants that differed only in how uncertain labels were handled during training and validation.\\
In the inclusive model, all labels originally marked as “uncertain” during label extraction (e.g., a report phrase such as “possible fracture”) were converted to “true” (finding present), thereby treating uncertain findings as positive cases. In the exclusive model, all “uncertain” labels were converted to “false” (finding absent), thereby treating uncertain findings as negative cases. Overall, the only difference between the two models was the handling of these “uncertain” labels in the training and validation datasets. All other parameters and architectures remained identical. \\
Definitive “true” and “false” labels were left unchanged in the training and validation sets, and all test sets contained only definitive labels. Each model was based on a modified ResNet50 architecture \cite{He.2015}: The original 1,000-class output layer was replaced with a new fully connected layer to output a vector of logits corresponding to each condition's binary classification, i.e., presence or absence. These logits were then passed through a sigmoid activation function to convert them into probabilities. For elbow and thumb datasets, which included both anteroposterior and lateral projections, each image was processed through separate ResNet50 networks with identity layers as final layers; their extracted feature vectors were concatenated before being fed into a final fully connected layer. \\
We used the AdamW optimizer initialized for training with a learning rate of 0.001. A step-based scheduler (StepLR) was employed to reduce the learning rate by 0.1 every seventh epoch.
Preprocessing included resizing the images to 512x512 pixels and augmentation with random flips, rotation, and color jitter.
The optimal decision threshold for the test set was determined on the validation set using the Youden index method \cite{YOUDEN.1950}. A summary of all training details can be found in Supplementary Table~\ref{tab:s1_training}.

\subsection{Statistical Analysis}
Statistical analyses were performed using R software (version 4.3.1; R Project for Statistical Computing) and Python. Label extraction performance was assessed using two metrics: (1) label-level accuracy, i.e., the proportion of labels correctly extracted across all reports, and (2) report-level accuracy, i.e., the proportion of reports where all labels were correctly extracted. Model performance metrics included accuracy, sensitivity, specificity, Area Under the Receiver Operating Characteristic Curve (AUC) and precision recall curves. Macro-averaged AUC values, i.e., the unweighted AUC averages computed for each label, were calculated across all labels with at least 10 positive cases in the internal and external test sets. This threshold was selected to strike a sensible balance between sufficient sample sizes for reliable AUC estimates and comprehensive multi-label evaluation. The DeLong test \cite{DeLong.1988} was used to compare AUC values between inclusive and exclusive models and between internal and external test sets, using the R pROC package \cite{Robin.2011}. Confidence intervals were determined by 1,000 bootstrap replicates, and the Benjamini–Hochberg correction \cite{Benjamini.1995} was applied to account for multiple comparisons against the significance threshold of p$<$0.05.

\section{Results}
\begin{figure}[H]
  \centering
  \includegraphics[width=0.7\linewidth]{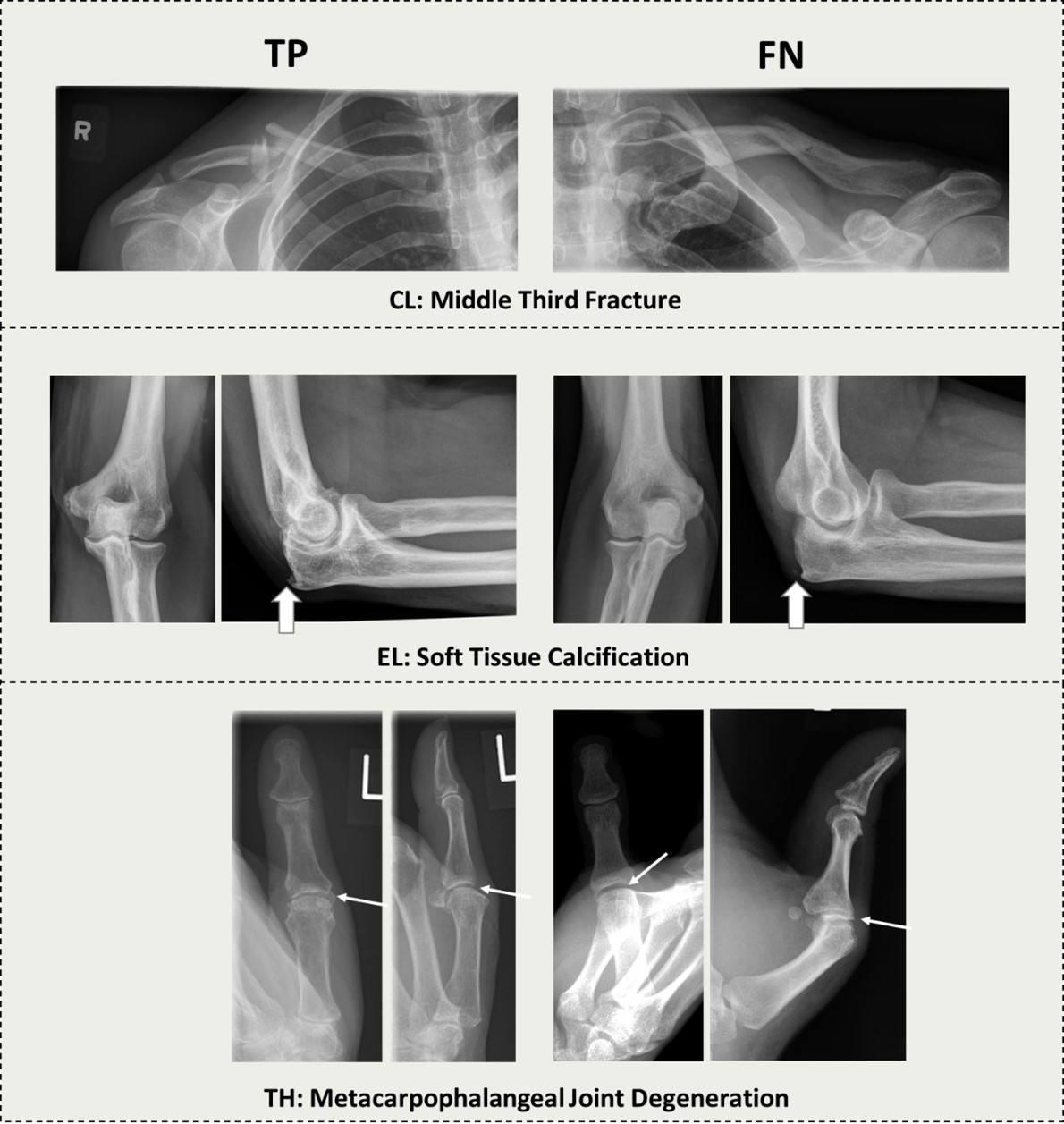}
  \caption{Representative Radiography Series Illustrating Model Performance Across Different Labels and Anatomic Regions. 
Shown are true positives (TP; left) and false negatives (FN; right) for the label indicated underneath each radiography series consisting of anteroposterior (left) and lateral projections (right). For the clavicle (CL), the model correctly identified a middle-third fracture in one patient (TP) and misclassified it as a medial third fracture in another patient (FN). For the elbow (EL), a soft tissue calcification at the triceps tendon insertion was accurately detected in one patient (TP) and missed in another (FN), likely because of its fainter and more subtle appearance. For the thumb (TH), metacarpophalangeal joint degeneration was correctly identified in one patient (TP) and missed in another patient (FN), likely because of a fixed Boutonnière-like deformity and consecutive superimposition of the metacarpus. Block arrows indicate calcifications, while arrows indicate signs of degeneration.
}
  \label{fig:examples}
\end{figure}

Patient demographics are detailed in Table~\ref{tab:demo}. Counts of conditions as a function of dataset, split, and model type are provided in Supplementary Tables ~\ref{tab:s2_clavicle_counts}, ~\ref{tab:s3_elbow_counts}, ~\ref{tab:s4_thumb_counts}. \\
\begin{table}[h]
\centering
\caption{Patient Demographics as a Function of Dataset (Internal, External), Split, and Anatomic Region. 
Patient age is presented as mean ± standard deviation. Patient sex is presented as count (\%).
}
\label{tab:demo}
\scriptsize
\begin{tabular}{l l l r l l}
\toprule
Dataset & Anatomic Region & Split & $N$ & Age [y] & Women \\
\midrule
\multirow{3}{*}{Internal} & \multirow{3}{*}{Clavicle} & Train & 749 & 46.4 $\pm$ 20.7 & 244 (32.6) \\
 & & Validation & 188 & 45.6 $\pm$ 20.8 & 47 (25.0) \\
 & & Test & 233 & 45.4 $\pm$ 19.8 & 80 (34.3) \\
\midrule
\multirow{3}{*}{Internal} & \multirow{3}{*}{Elbow} & Train & 2408 & 48.4 $\pm$ 20.8 & 1031 (42.9) \\
 & & Validation & 602 & 49.0 $\pm$ 20.3 & 271 (45.2) \\
 & & Test & 745 & 48.5 $\pm$ 20.9 & 305 (40.9) \\
\midrule
\multirow{3}{*}{Internal} & \multirow{3}{*}{Thumb} & Train & 1268 & 41.6 $\pm$ 17.4 & 482 (38.0) \\
 & & Validation & 317 & 41.1 $\pm$ 17.6 & 132 (41.6) \\
 & & Test & 393 & 41.9 $\pm$ 18.2 & 158 (40.2) \\
\midrule
External & Clavicle & Test & 300 & 47.3 $\pm$ 18.5 & 113 (37.7) \\
External & Elbow & Test & 300 & 45.5 $\pm$ 17.8 & 127 (42.3) \\
External & Thumb & Test & 300 & 41.7 $\pm$ 16.0 & 132 (44.0) \\
\bottomrule
\end{tabular}
\end{table}
The label extraction process functioned as intended, and the tool successfully filled out the templates using the specified JSON formats. For the internal data, clavicle reports achieved 98.8\% label-level accuracy (5,988/6,058) and 78.5\% report-level accuracy (183/233 reports with 26 labels each). Elbow reports reached 98.6\% (21,294/21,605) and 74.4\% (554/745, 29 labels per report), whereas thumb reports achieved 99.0\% (9,731/9,825) and 85.5\% (336/393, 25 labels per report).\\
In the external test set of 300 radiologic reports per joint, label extraction performance was similarly strong: For the clavicle, label-level accuracy was 98.6\% (7,687/7,800) and report-level accuracy was 72.7\% (218/300). For the elbow, the accuracy rates were 98.4\% (8,564/8,700) and 71.3 \% (214/300), and for the thumb, they were 98.1\% (7,354/7,500) and 73.7\% (221/300).\\
Detecting uncertain labels posed a challenge. For the clavicle, manual identification found uncertain labels in 3.9\% (9/233) of reports, but only two (0.9\%) were automatically detected. For the elbow, uncertain labels were present in 10.5\% (78/745) of reports, with the tool identifying 48 (6.4\%). For the thumb, uncertain labels were present in 9.7\% (38/393), with the tool identifying 21 (5.3\%). A similar trend was observed for the external test set: Uncertainty was present in 5.3\% (16/300; clavicle), 16.3\% (49/300; elbow), and 16.0\% (48/300; thumb), respectively, of which the extraction pipeline detected 10 (3.3\%), 26 (8.7\%), and 40 (13.3\%).
The training and validation sets contained few such labels—42 for clavicle, 492 for elbow, and 231 for thumb (Supplementary Table \ref{tab:s5_tfu_counts}).\\
The models achieved competitive performance across the anatomic regions, but per-label performance varied. Figure ~\ref{fig:examples} illustrates representative true positive and false negative example images.
Tables \ref{tab:clavicle}, \ref{tab:elbow} and \ref{tab:thumb} detail AUC values (for labels with n$\geq10$) and p-values of the DeLong test, while additional performance metrics, i.e., accuracy, sensitivity, and specificity as well as ROC and precision recall curves, can be found in the supplement (Supplementary Tables ~\ref{tab:s6_clavicle_internal},~\ref{tab:s7_clavicle_external},~\ref{tab:s8_elbow_internal},~\ref{tab:s9_elbow_external},~\ref{tab:s10_thumb_internal},~\ref{tab:s11_thumb_external} and Supplementary Figures ~\ref{fig:s1_clavicle_roc},~\ref{fig:s1_elbow_roc},~\ref{fig:s1_thumb_roc}). In brief, these precision recall curves reflected the impact of class imbalance, with rare labels (e.g., “Ossicles” [elbow and thumb]) showing steep early precision drop-offs despite moderate AUC values, whereas high-prevalence labels (e.g., “Fracture [All Locations]” [elbow and thumb]) maintained high precision across a broader recall range. Across all regions, both model variants tended to demonstrate high specificity for rare findings but reduced sensitivity, particularly for subtle or soft tissue–related abnormalities.\\

For the clavicle, the macro-averaged AUC was 0.80 (range, 0.59 to 0.95) for the inclusive and 0.81 (range, 0.63 to 0.94) for the exclusive model (p$\geq$0.49). We observed high AUC values for common findings —labels with many positive cases— such as “Fracture (All Locations).” In contrast, less common findings had moderate AUC values, such as specific fracture locations like “Middle Third Fracture,” and performance was poorer for rare findings like “Acromioclavicular Joint–Joint Space Widened.” Across labels, only minimal differences in AUC values were found between the inclusive and exclusive models, indicating that the handling of uncertain labels did not significantly affect model performance. The models maintained good performance on common findings in the external dataset, while challenging labels (e.g., “Swelling or Hematoma”) had weak performance on both internal and external datasets.\\
For the elbow, the macro-averaged AUC was 0.80 (range, 0.62 to 0.87) for the inclusive and 0.80 for the exclusive model (range, 0.61 to 0.88) (p$\geq$0.47). Again, common fracture-related findings (e.g., “Radial Head Fracture”) had high AUC values, while less prevalent findings such as “Ossicles” had lower and less consistent AUC values in the internal and external test sets. Model type and dataset did not significantly impact performance metrics.\\
For the thumb, the macro-averaged AUC was 0.76 (range, 0.59 to 0.91) for the inclusive and 0.78 for the exclusive model (range, 0.61 to 0.90) (p$\geq0.65$). AUC values were highest for joint degeneration-associated labels, both for the carpometacarpal and metacarpophalangeal articulations, while detecting fractures and soft tissue changes such as “Swelling/Dactylitis” was considerably less efficient, regardless of model type or dataset.\\
While no significant changes in AUC could be observed when comparing model type and dataset source, other metrics occasionally showed larger swings: For some, sensitivity increased while specificity decreased, or vice-versa (e.g. “Displacement” in the clavicle external set shows an 80\% vs 70\% sensitivity shift that is offset by the opposite change in specificity). However, these fluctuations are tied to the single operating point imposed by Youden’s index. When the threshold-independent AUC was compared with the DeLong test, no label showed a significant inclusive–exclusive or internal–external difference. 

\begin{table}[H]
\centering
\caption{Performance of the Image Classification Models Trained with Automatically Extracted Labels for Radiographs of the Clavicle.}
\label{tab:clavicle}
\scriptsize
\begin{tabular}{l l r l l l}
\toprule
Label & Dataset & $N$ & Inclusive Model & Exclusive Model & $p$-value [$^{**}$] \\
\midrule

\multirow{3}{*}{Fracture (All Locations)}
 & Internal & 121 & 0.95 (0.91, 0.97) & 0.94 (0.90, 0.96) & 0.81 \\
 & External & 147 & 0.91 (0.88, 0.94) & 0.91 (0.88, 0.95) & 0.95 \\
 & \textit{p-value [$*$]} &  & 0.73 & 0.73 &  \\
\midrule

\multirow{3}{*}{Displacement}
 & Internal & 98 & 0.91 (0.87, 0.94) & 0.90 (0.85, 0.94) & 0.91 \\
 & External & 123 & 0.93 (0.90, 0.95) & 0.92 (0.88, 0.95) & 0.81 \\
 & \textit{p-value [$*$]} &  & 0.73 & 0.89 &  \\
\midrule

\multirow{3}{*}{Comminuted or Fragmented Fracture (All Locations)}
 & Internal & 43 & 0.89 (0.84, 0.93) & 0.89 (0.84, 0.93) & 0.95 \\
 & External & 59 & 0.91 (0.87, 0.94) & 0.91 (0.87, 0.94) & 0.95 \\
 & \textit{p-value [$*$]} &  & 0.76 & 0.73 &  \\
\midrule

\multirow{3}{*}{Middle Third Fracture}
 & Internal & 66 & 0.84 (0.78, 0.89) & 0.86 (0.80, 0.91) & 0.81 \\
 & External & 76 & 0.90 (0.86, 0.94) & 0.88 (0.84, 0.92) & 0.64 \\
 & \textit{p-value [$*$]} &  & 0.73 & 0.73 &  \\
\midrule

\multirow{3}{*}{Lateral Third Fracture}
 & Internal & 45 & 0.83 (0.75, 0.89) & 0.83 (0.76, 0.90) & 0.95 \\
 & External & 67 & 0.81 (0.75, 0.87) & 0.83 (0.77, 0.88) & 0.81 \\
 & \textit{p-value [$*$]} &  & 0.91 & 0.91 &  \\
\midrule

\multirow{3}{*}{Acromioclavicular Joint Degeneration}
 & Internal & 18 & 0.78 (0.65, 0.89) & 0.77 (0.64, 0.87) & 0.91 \\
 & External & 28 & 0.74 (0.65, 0.82) & 0.74 (0.65, 0.82) & 0.95 \\
 & \textit{p-value [$*$]} &  & 0.91 & 0.91 &  \\
\midrule

\multirow{3}{*}{Joint Degeneration (All Locations)}
 & Internal & 19 & 0.78 (0.65, 0.90) & 0.76 (0.64, 0.87) & 0.81 \\
 & External & 59 & 0.75 (0.68, 0.82) & 0.76 (0.68, 0.83) & 0.95 \\
 & \textit{p-value [$*$]} &  & 0.91 & 0.95 &  \\
\midrule

\multirow{3}{*}{Acromioclavicular Joint -- Joint Space Widened}
 & Internal & 14 & 0.60 (0.46, 0.72) & 0.63 (0.47, 0.77) & 0.86 \\
 & External & 34 & 0.53 (0.42, 0.64) & 0.51 (0.41, 0.62) & 0.91 \\
 & \textit{p-value [$*$]} &  & 0.81 & 0.81 &  \\
\midrule

\multirow{3}{*}{Swelling or Hematoma}
 & Internal & 12 & 0.59 (0.39, 0.75) & 0.70 (0.55, 0.83) & 0.49 \\
 & External & 34 & 0.47 (0.36, 0.58) & 0.54 (0.42, 0.65) & 0.46 \\
 & \textit{p-value [$*$]} &  & 0.81 & 0.64 &  \\
\bottomrule
\end{tabular}
\vspace{2pt}
\raggedright \footnotesize \textit{Note:} Area under the receiver operating characteristic curve (AUC) values are shown with 95\% confidence intervals in parentheses for each label with at least 10 positive cases in both internal and external test sets. N refers to the number of positive cases in the corresponding test set, which was identical for the inclusive and exclusive models. p-values from the DeLong test compare (i) internal vs. external test sets for the same model (*, row-wise) and (ii) inclusive vs. exclusive models within the same test set (**, column-wise).
\end{table}

\begin{table}[H]
\centering
\caption{Performance of the Image Classification Models Trained with Automatically Extracted Labels for Radiographs of the Elbow.}
\label{tab:elbow}
\scriptsize
\begin{tabular}{l l r l l l}
\toprule
Label & Dataset & $N$ & Inclusive Model & Exclusive Model & $p$-value [$^{**}$] \\
\midrule

\multirow{3}{*}{Fracture (All Locations)}
 & Internal & 162 & 0.87 (0.83, 0.90) & 0.88 (0.85, 0.91) & 0.86 \\
 & External & 56 & 0.85 (0.79, 0.90) & 0.85 (0.79, 0.90) & 0.99 \\
 & \textit{p-value [$*$]} &  & 0.86 & 0.86 &  \\
\midrule

\multirow{3}{*}{Radius Fracture}
 & Internal & 112 & 0.86 (0.82, 0.90) & 0.85 (0.82, 0.89) & 0.86 \\
 & External & 44 & 0.84 (0.76, 0.91) & 0.88 (0.81, 0.94) & 0.41 \\
 & \textit{p-value [$*$]} &  & 0.92 & 0.86 &  \\
\midrule

\multirow{3}{*}{Radial Head Fracture}
 & Internal & 102 & 0.86 (0.81, 0.90) & 0.85 (0.81, 0.89) & 0.86 \\
 & External & 42 & 0.87 (0.80, 0.93) & 0.89 (0.84, 0.94) & 0.73 \\
 & \textit{p-value [$*$]} &  & 0.89 & 0.73 &  \\
\midrule

\multirow{3}{*}{Exostosis}
 & Internal & 57 & 0.82 (0.76, 0.87) & 0.81 (0.76, 0.86) & 0.86 \\
 & External & 40 & 0.86 (0.80, 0.91) & 0.85 (0.78, 0.91) & 0.86 \\
 & \textit{p-value [$*$]} &  & 0.86 & 0.86 &  \\
\midrule

\multirow{3}{*}{Joint Degeneration (All Locations)}
 & Internal & 52 & 0.81 (0.74, 0.86) & 0.82 (0.75, 0.87) & 0.86 \\
 & External & 39 & 0.84 (0.77, 0.92) & 0.84 (0.77, 0.91) & 0.99 \\
 & \textit{p-value [$*$]} &  & 0.86 & 0.86 &  \\
\midrule

\multirow{3}{*}{Sclerotic Lesion}
 & Internal & 12 & 0.80 (0.66, 0.92) & 0.85 (0.74, 0.94) & 0.47 \\
 & External & 21 & 0.65 (0.53, 0.77) & 0.63 (0.52, 0.75) & 0.86 \\
 & \textit{p-value [$*$]} &  & 0.54 & 0.15 &  \\
\midrule

\multirow{3}{*}{Fat Pad Sign}
 & Internal & 95 & 0.78 (0.72, 0.84) & 0.78 (0.72, 0.84) & 0.99 \\
 & External & 38 & 0.78 (0.68, 0.86) & 0.78 (0.69, 0.86) & 0.99 \\
 & \textit{p-value [$*$]} &  & 0.99 & 0.99 &  \\
\midrule

\multirow{3}{*}{Soft Tissue Calcifications}
 & Internal & 61 & 0.78 (0.71, 0.83) & 0.76 (0.68, 0.83) & 0.73 \\
 & External & 30 & 0.61 (0.49, 0.71) & 0.63 (0.52, 0.73) & 0.86 \\
 & \textit{p-value [$*$]} &  & 0.23 & 0.41 &  \\
\midrule

\multirow{3}{*}{Ossicles}
 & Internal & 19 & 0.62 (0.48, 0.74) & 0.61 (0.48, 0.72) & 0.99 \\
 & External & 16 & 0.78 (0.66, 0.90) & 0.75 (0.60, 0.87) & 0.86 \\
 & \textit{p-value [$*$]} &  & 0.41 & 0.54 &  \\ \\
\bottomrule
\end{tabular}
\vspace{2pt}
\raggedright \footnotesize \textit{\\Note:} Area under the receiver operating characteristic curve (AUC) values are shown with 95\% confidence intervals in parentheses for each label with at least 10 positive cases in both internal and external test sets. N refers to the number of positive cases in the corresponding test set, which was identical for the inclusive and exclusive models. p-values from the DeLong test compare (i) internal vs. external test sets for the same model (*, row-wise) and (ii) inclusive vs. exclusive models within the same test set (**, column-wise).
\end{table}

\begin{table}[H]
\centering
\caption{Performance of the Image Classification Models Trained with Automatically Extracted Labels for Radiographs of the Thumb.}
\label{tab:thumb}
\scriptsize
\begin{tabular}{l l r l l l}
\toprule
Label & Dataset & $N$ & Inclusive Model & Exclusive Model & $p$-value [$^{**}$] \\
\midrule

\multirow{3}{*}{Carpometacarpal Joint Degeneration}
 & Internal & 30 & 0.91 (0.85, 0.96) & 0.90 (0.85, 0.95) & 0.95 \\
 & External & 30 & 0.90 (0.84, 0.95) & 0.90 (0.85, 0.95) & 0.96 \\
 & \textit{p-value [$*$]} &  & 0.95 & 0.96 &  \\
\midrule

\multirow{3}{*}{Joint Degeneration (All Locations)}
 & Internal & 52 & 0.89 (0.84, 0.93) & 0.88 (0.81, 0.93) & 0.69 \\
 & External & 49 & 0.90 (0.85, 0.95) & 0.91 (0.86, 0.95) & 0.73 \\
 & \textit{p-value [$*$]} &  & 0.95 & 0.71 &  \\
\midrule

\multirow{3}{*}{Metacarpophalangeal Joint Degeneration}
 & Internal & 21 & 0.88 (0.78, 0.95) & 0.89 (0.79, 0.96) & 0.73 \\
 & External & 20 & 0.85 (0.75, 0.92) & 0.85 (0.77, 0.93) & 1 \\
 & \textit{p-value [$*$]} &  & 0.72 & 0.54 &  \\
\midrule

\multirow{3}{*}{Distal Phalanx -- Comminuted or Fragmented Fracture}
 & Internal & 14 & 0.85 (0.69, 0.97) & 0.89 (0.74, 0.99) & 0.65 \\
 & External & 10 & 0.66 (0.47, 0.82) & 0.72 (0.58, 0.84) & 0.69 \\
 & \textit{p-value [$*$]} &  & 0.65 & 0.65 &  \\
\midrule

\multirow{3}{*}{Distal Phalanx Fracture}
 & Internal & 49 & 0.75 (0.66, 0.84) & 0.77 (0.69, 0.85) & 0.73 \\
 & External & 28 & 0.67 (0.58, 0.76) & 0.68 (0.58, 0.79) & 0.95 \\
 & \textit{p-value [$*$]} &  & 0.65 & 0.65 &  \\
\midrule

\multirow{3}{*}{Joint Subluxation (All Locations)}
 & Internal & 18 & 0.72 (0.57, 0.85) & 0.69 (0.56, 0.82) & 0.65 \\
 & External & 16 & 0.71 (0.53, 0.86) & 0.70 (0.53, 0.88) & 0.96 \\
 & \textit{p-value [$*$]} &  & 0.96 & 0.98 &  \\
\midrule

\multirow{3}{*}{Fracture (All Locations)}
 & Internal & 73 & 0.70 (0.63, 0.77) & 0.71 (0.64, 0.78) & 0.95 \\
 & External & 55 & 0.62 (0.54, 0.70) & 0.65 (0.57, 0.73) & 0.67 \\
 & \textit{p-value [$*$]} &  & 0.65 & 0.67 &  \\
\midrule

\multirow{3}{*}{Swelling/Dactylitis}
 & Internal & 23 & 0.68 (0.55, 0.81) & 0.73 (0.63, 0.81) & 0.67 \\
 & External & 47 & 0.57 (0.50, 0.66) & 0.60 (0.51, 0.69) & 0.71 \\
 & \textit{p-value [$*$]} &  & 0.65 & 0.65 &  \\
\midrule

\multirow{3}{*}{Ossicles}
 & Internal & 32 & 0.64 (0.53, 0.73) & 0.66 (0.56, 0.76) & 0.65 \\
 & External & 36 & 0.42 (0.33, 0.53) & 0.46 (0.36, 0.56) & 0.67 \\
 & \textit{p-value [$*$]} &  & 0.17 & 0.18 &  \\
\midrule

\multirow{3}{*}{Proximal Phalanx Fracture}
 & Internal & 16 & 0.61 (0.47, 0.74) & 0.65 (0.53, 0.77) & 0.71 \\
 & External & 23 & 0.46 (0.33, 0.58) & 0.53 (0.41, 0.65) & 0.65 \\
 & \textit{p-value [$*$]} &  & 0.65 & 0.65 &  \\
\midrule

\multirow{3}{*}{Metacarpophalangeal Joint Subluxation}
 & Internal & 11 & 0.59 (0.41, 0.75) & 0.61 (0.39, 0.82) & 0.95 \\
 & External & 10 & 0.64 (0.44, 0.83) & 0.61 (0.36, 0.85) & 0.96 \\
 & \textit{p-value [$*$]} &  & 0.95 & 1 \\
\bottomrule
\end{tabular}
\vspace{2pt}
\raggedright \footnotesize \textit{Note:} Area under the receiver operating characteristic curve (AUC) values are shown with 95\% confidence intervals in parentheses for each label with at least 10 positive cases in both internal and external test sets. N refers to the number of positive cases in the corresponding test set, which was identical for the inclusive and exclusive models. p-values from the DeLong test compare (i) internal vs. external test sets for the same model (*, row-wise) and (ii) inclusive vs. exclusive models within the same test set (**, column-wise).
\end{table}

\section{Discussion}
This study used GPT-4o for automatic label extraction from free-text radiologic reports of three anatomic regions of the upper extremity: radiography of the clavicle, elbow, and thumb. The LLM extracted labels with an accuracy ranging from 98.1\% to 99.0\%, which were used to train out-of-the-box multi-label classification models with competitive and robust diagnostic performance. However, detecting uncertain labels was less successful, with the LLM identifying only a portion of the uncertain cases identified manually and label uncertainty did not significantly impact model performance. \\
Automatic label extraction allowed us to assemble training data rapidly and to build reliable classifiers for the frequent, bone-related labels:  AUC values for “fracture,” “displacement,” and similar findings consistently exceeded 0.90. In contrast, performance on soft-tissue abnormalities (e.g. “swelling”, ”soft tissue masses”) was markedly lower. These labels were both rarer and intrinsically harder to identify on radiographs. The limited number of positive cases, combined with greater subjectivity in the ground-truth reports, provided the network with a weaker and noisier learning signal. As a result, confidence intervals widened, sensitivity and specificity fluctuated, and no statistically significant AUC differences emerged between the inclusive and exclusive uncertainty-handling schemes.\\
Our results confirm previous studies investigating automatic labeling approaches. One original archetypal approach is the CheXpert labeler, a rule-based NLP tool used for automatic label extraction of chest radiography reports and subsequent training of classification models \cite{Irvin.2019}. While the CheXpert labeler and similar models (e.g., MIMIC-XCR labeler \cite{Johnson.2019}) provide an efficient means to annotate large datasets, they are challenged by complex language, variable reporting styles, and negations and uncertainties, resulting in label inaccuracies of up to 10\% \cite{Sun.2022},\cite{Irvin.2019}. Recently, LLM-based methods for label extraction have been demonstrated to be more accurate due to their superior understanding of natural language \cite{Gu.2024}. Al Mohamed et al. used an open-weight LLM to extract a single binary label, i.e., fracture present or not, from reports of ankle radiographs \cite{AlMohamad.2025}. Their report-level accuracy of 92\% was considerably higher than our study’s multi-label report-level accuracy of 74\% to 86\%. Their classification network trained on this label reached an AUC value of 0.93, comparable to our performance for some fracture labels. However, their study was limited to a single anatomic region and did not address label uncertainty. \\
Comparing our fracture-classification results to the meta-review in \cite{Kuo.2022}, we note that the pooled AUC of 0.97 reported for classical fracture-detection CNNs—slightly above the 0.95 we achieve for clavicle fractures—was obtained from hand-curated datasets and models that focus on single or few labels only. To our knowledge, no published work has attempted a comprehensive multi-label prediction task for clavicle, elbow or thumb and few studies of any kind address these regions because they are imaged far less frequently than, for example, the chest. The main added value of our study is therefore methodological: by combining GPT-4o label extraction with routine reports we show that competitive classifiers can be trained rapidly even for sparsely imaged anatomies and for a spectrum of findings, rather than for fractures alone. \\
Handling uncertainty proved challenging. Although we aimed to standardize the uncertainty terms based on pertinent literature \cite{Callen.2020} and provided examples in the prompt, the LLM identified fewer uncertainty terms than were manually identified. Overall, the occurrence of uncertain labels in the ground truth was surprisingly low compared to other studies \cite{Callen.2020}. The interpretation of uncertainty is shaped by personal preference and experience, institutional policies, and healthcare system-related circumstances. For example, if the medicolegal environment is more litigious, as in the United States, uncertainty terms and hedging language may be more prevalent due to fear of malpractice litigation \cite{Baungaard.2022}. However, we found that uncertainty handling did not significantly impact model performance. In line with this observation, Irivin et al. reported that converting uncertain labels to positive versus negative labels did not significantly impact diagnostic performance, except for atelectasis \cite{Irvin.2019}. Because only a small fraction of labels was flagged as uncertain, the absence of performance differences between the inclusive and exclusive models may simply reflect limited statistical power. Nevertheless, the finding also suggests that our classifiers tolerate a modest level of label uncertainty without a measurable influence on performance.\\
This study has limitations. First, despite our intention to conduct a large-scale analysis, overall patient numbers were relatively low because of strict inclusion criteria. Data availability was insufficient for some labels, and we had to exclude those labels with fewer than 10 cases in the test sets. Even then, data scarcity for some labels led to wide confidence intervals and might have contributed to insignificant Delong test results. Second, we included German language reports; hence, whether the results hold when using less common languages is questionable. Fourth, the model did not provide visual outputs for model explainability. Fifth, model performance after training using automatically extracted labels was not compared to training using human reference labels, so the effect of LLM-based mislabeling remains unaccounted for. Sixth, the inherent non-deterministic nature of LLM outputs makes reproducing the results difficult.\\
Therefore, future research should prioritize multi-institutional collaborations to include multi-label datasets for rare and subtle findings across more anatomic regions, conditions, modalities, and radiologic reports in different languages.\\
In conclusion, this study demonstrates that state-of-the-art large language models effectively extract labels from radiologic reports with high accuracy and can be utilized to train multi-label classification models. While detecting label uncertainty remains challenging, its impact on model performance is minimal. Clinical data not originally intended for model development can thus be converted into structured formats suitable for expedited and decentralized AI model development and/or fine-tuning.

\section*{Code Availability}
Code for label extraction and model training: \url{https://github.com/TruhnLab/GPT_Label_Extraction_MSK}.

\section*{Acknowledgements}
This research is supported by the Deutsche Forschungsgemeinschaft - DFG (701010997, 517243167, 515639690) , the German Federal Ministry of Research, Technology and Space (Transform Liver - 031L0312C, DECIPHER-M, 01KD2420B) and the European Union Research and Innovation Programme (ODELIA - GA 101057091, SAGMA – GA 101222556). The data used in this publication was managed using the research data management platform Coscine (http://doi.org/10.17616/R31NJNJZ) with storage space of the Research Data Storage (RDS) (DFG: INST222/1261-1) and DataStorage.nrw (DFG: INST222/1530-1) granted by the DFG and Ministry of Culture and Science of the State of North Rhine-Westphalia. The authors gratefully acknowledge the computing time provided to them at the NHR Center NHR4CES at RWTH Aachen University (project number p0021834). This is funded by the Federal Ministry of Research, Technology and Space , and the state governments participating on the basis of the resolutions of the GWK for national high performance computing at universities (www.nhr-verein.de/unsere-partner).

\section*{Competing Interests}
D.T. received honoraria for lectures by Bayer, GE, Roche, Astra Zenica, and Philips and holds shares in StratifAI GmbH,
11/15
Germany and in Synagen GmbH, Germany

\bibliographystyle{unsrt}
\bibliography{references}

\clearpage
\section*{Supplementary Material}
% ---- Supplementary numbering ----
\renewcommand{\thetable}{S\arabic{table}}
\setcounter{table}{0}
% ---- Supplementary Figure numbering ----
\renewcommand{\thefigure}{S\arabic{figure}}
\setcounter{figure}{0}

\newcounter{supptext}
\renewcommand{\thesupptext}{\arabic{supptext}} % 1,2,3...
\section*{Supplementary Text 1: Prompts for Automated Label Extraction}
\phantomsection
\refstepcounter{supptext}  
\label{suppl_text_1}
\subsection*{Clavicle}
\textit{Instruction to the model:}\\
Fill out the following template \verb|{template_json}| in JSON format according to the information given in \verb|{finding}|. Adhere strictly to the structure of the template. Only fill out the \verb|"finding"| fields and do not add anything else.
\begin{enumerate}
\item Mark a finding as \textbf{true} only if it is explicitly confirmed in the report with no uncertainty or hedging terms.
\item Mark a finding as \textbf{uncertain} if the report contains hedging or uncertainty terms like ``not clearly delineable'' (German: ``nicht sicher abgrenzbar''), ``most likely'' (``am ehesten''), ``possibly'' (``m{\"o}glicherweise''), ``cannot be ruled out'' (``nicht ausgeschlossen''), ``suspected'' (``Verdacht auf''), ``a potential differential diagnosis may be X'' (``als DD k{\"a}me in Frage X'' or ``DD X'') or similar phrases indicating doubt.
\item Mark a finding as \textbf{false} if the report explicitly states the absence of a finding or if the report contains no information related to the specific label.
\item If a specific sub-category finding (e.g., ``Lateral Third Fracture'') is marked as true or uncertain, ensure that the broader category (e.g., ``Fracture [All Locations]'') is marked accordingly.
\item Ensure no findings from other anatomic regions are considered when marking the \verb|"finding"| fields. Each finding must be strictly limited to the anatomic region of interest (e.g., clavicle) specified in the template.
\end{enumerate}

\subsection*{Elbow}
\textit{Instruction to the model:}\\
Fill out the following template \verb|{template_json}| in JSON format according to the information given in \verb|{finding}|. Adhere strictly to the structure of the template. Only fill out the \verb|"finding"| fields and do not add anything else.
\begin{enumerate}
\item Mark a finding as \textbf{true} only if it is explicitly confirmed in the report with no uncertainty or hedging terms.
\item Mark a finding as \textbf{uncertain} if the report contains hedging or uncertainty terms like ``not clearly delineable'' (German: ``nicht sicher abgrenzbar''), ``most likely'' (``am ehesten''), ``possibly'' (``m{\"o}glicherweise''), ``cannot be ruled out'' (``nicht ausgeschlossen''), ``suspected'' (``Verdacht auf''), ``a potential differential diagnosis may be X'' (``als DD k{\"a}me in Frage X'' or ``DD X'') or similar phrases indicating doubt.
\item Mark a finding as \textbf{false} if the report explicitly states the absence of a finding or if the report contains no information related to the specific label.
\item If a specific sub-category finding (e.g., ``Radial Head Fracture'') is marked as true or uncertain, ensure that the broader category (e.g., ``Fracture [All Locations]'') is marked accordingly.
\item Ensure no findings from other anatomic regions are considered when marking the \verb|"finding"| fields. Each finding must be strictly limited to the anatomic region of interest (e.g., elbow) specified in the template.
\end{enumerate}

\subsection*{Thumb}
\textit{Instruction to the model:}\\
Fill out the following template \verb|{template_json}| in JSON format according to the information given in \verb|{finding}|. Adhere strictly to the structure of the template. Only fill out the \verb|"finding"| fields and do not add anything else.
\begin{enumerate}
\item Mark a finding as \textbf{true} only if it is explicitly confirmed in the report with no uncertainty or hedging terms.
\item Mark a finding as \textbf{uncertain} if the report contains hedging or uncertainty terms like ``not clearly delineable'' (German: ``nicht sicher abgrenzbar''), ``most likely'' (``am ehesten''), ``possibly'' (``m{\"o}glicherweise''), ``cannot be ruled out'' (``nicht ausgeschlossen''), ``suspected'' (``Verdacht auf''), ``a potential differential diagnosis may be X'' (``als DD k{\"a}me in Frage X'' or ``DD X'') or similar phrases indicating doubt.
\item Mark a finding as \textbf{false} if the report explicitly states the absence of a finding or if the report contains no information related to the specific label.
\item If a specific sub-category finding (e.g., ``First Metacarpal Bone Fracture'') is marked as true or uncertain, ensure that the broader category (e.g., ``Fracture [All Locations]'') is marked accordingly.
\item Ensure no findings from other anatomic regions are considered when marking the \verb|"finding"| fields. Each finding must be strictly limited to the anatomic region of interest (e.g., thumb) specified in the template.
\end{enumerate}

\section*{Supplementary Text 2: Templates for Automated Label Extraction}
\phantomsection
\refstepcounter{supptext} 
\label{suppl_text_2}
For each label, the tool was instructed to choose one of: \texttt{true} (present), \texttt{false} (absent; default), or \texttt{uncertain} (hedging/doubt in the report).

\subsection*{Template for the Clavicle}
\begin{verbatim}
{
  "Fracture (All Locations)": {"finding": false},
  "Medial Third Fracture": {"finding": false},
  "Middle Third Fracture": {"finding": false},
  "Lateral Third Fracture": {"finding": false},
  "Comminuted or Fragmented Fracture (All Locations)": {"finding": false},
  "Displacement": {"finding": false},
  "Sclerotic Lesion": {"finding": false},
  "Lytic Lesion": {"finding": false},
  "Joint Dislocation (All Locations)": {"finding": false},
  "Joint Subluxation (All Locations)": {"finding": false},
  "Joint Degeneration (All Locations)": {"finding": false},
  "Acromioclavicular Joint - Joint Space widened": {"finding": false},
  "Acromioclavicular Joint - Joint Space narrowed": {"finding": false},
  "Acromioclavicular Joint - Subluxation": {"finding": false},
  "Acromioclavicular Joint - Dislocation": {"finding": false},
  "Acromioclavicular Joint Degeneration": {"finding": false},
  "Sternoclavicular Joint - Joint Space widened": {"finding": false},
  "Sternoclavicular Joint - Joint Space narrowed": {"finding": false},
  "Sternoclavicular Joint - Subluxation": {"finding": false},
  "Sternoclavicular Joint - Dislocation": {"finding": false},
  "Sternoclavicular Joint Degeneration": {"finding": false},
  "Swelling or Hematoma": {"finding": false},
  "Soft Tissue Calcifications": {"finding": false},
  "Soft Tissues Masses or Mass-like lesions": {"finding": false},
  "Foreign Bodies": {"finding": false},
  "Ossicles": {"finding": false}
}
\end{verbatim}

\subsection*{Template for the Elbow}
\begin{verbatim}
{
  "Fracture (All Locations)": {"finding": false},
  "Lytic Lesion": {"finding": false},
  "Sclerotic Lesion": {"finding": false},
  "Distal Humerus - Fracture": {"finding": false},
  "Distal Humerus - Comminuted or Fragmented Fracture": {"finding": false},
  "Distal Humerus - Displacement": {"finding": false},
  "Distal Humerus Fracture - Extension into Joint": {"finding": false},
  "Olecranon Fracture": {"finding": false},
  "Olecranon - Displaced Fracture": {"finding": false},
  "Olecranon - Comminuted or Fragmented Fracture": {"finding": false},
  "Coronoid Process Fracture": {"finding": false},
  "Coronoid Process - Avulsion of the tip": {"finding": false},
  "Ulna Fracture": {"finding": false},
  "Radial Head Fracture": {"finding": false},
  "Radial Head - Displaced": {"finding": false},
  "Radial Head - Comminuted or Fragmented Fracture": {"finding": false},
  "Radial Neck Fracture": {"finding": false},
  "Radial Neck - Displaced": {"finding": false},
  "Radial Neck - Comminuted or Fragmented Fracture": {"finding": false},
  "Radius Fracture": {"finding": false},
  "Joint Subluxation (All Locations)": {"finding": false},
  "Joint Dislocation (All Locations)": {"finding": false},
  "Joint Degeneration (All Locations)": {"finding": false},
  "Soft Tissue Calcifications": {"finding": false},
  "Soft Tissue Masses or Mass-like lesions": {"finding": false},
  "Fat Pad Sign": {"finding": false},
  "Foreign Bodies": {"finding": false},
  "Ossicles": {"finding": false},
  "Exostosis": {"finding": false}
}
\end{verbatim}

\subsection*{Template for the Thumb}
\begin{verbatim}
{
  "Fracture (All Locations)": {"finding": false},
  "Comminuted or Fragmented Fracture (All Locations)": {"finding": false},
  "First Metacarpal Bone Fracture": {"finding": false},
  "First Metacarpal Bone - Comminuted or Fragmented Fracture": {"finding": false},
  "Proximal Phalanx Fracture": {"finding": false},
  "Proximal Phalanx - Comminuted or Fragmented Fracture": {"finding": false},
  "Distal Phalanx Fracture": {"finding": false},
  "Distal Phalanx - Comminuted or Fragmented Fracture": {"finding": false},
  "Joint Subluxation (All Locations)": {"finding": false},
  "Joint Dislocation (All Locations)": {"finding": false},
  "Joint Degeneration (All Locations)": {"finding": false},
  "Carpometacarpal Joint - Subluxation": {"finding": false},
  "Carpometacarpal Joint - Dislocation": {"finding": false},
  "Carpometacarpal Joint Degeneration": {"finding": false},
  "Metacarpophalangeal Joint - Subluxation": {"finding": false},
  "Metacarpophalangeal Joint - Dislocation": {"finding": false},
  "Metacarpophalangeal Joint Degeneration": {"finding": false},
  "Interphalangeal Joint - Subluxation": {"finding": false},
  "Interphalangeal Joint - Dislocation": {"finding": false},
  "Interphalangeal Joint Degeneration": {"finding": false},
  "Swelling/Dactylitis": {"finding": false},
  "Soft Tissue Calcifications": {"finding": false},
  "Soft Tissues Masses or Mass-like lesions": {"finding": false},
  "Foreign Bodies": {"finding": false},
  "Ossicles": {"finding": false}
}
\end{verbatim}

\begin{table}[h]
\caption{Training Details for the Classification Models.}
\label{tab:s1_training}
\centering
\setlength{\tabcolsep}{6pt} % optional: adjust column spacing
\renewcommand{\arraystretch}{1.1} % optional: adjust row spacing
\begin{tabular}{@{} p{0.35\textwidth} p{0.55\textwidth} @{}} % two wrapping columns
\toprule
\textbf{Category} & \textbf{Description / Value}\\
\midrule
Framework & PyTorch 2.4.0\\
Pre-trained backbone & ResNet-50 (ImageNet-1k weights)\\
Input size & $512\times512$ pixels\\
Normalization & mean [0.485, 0.456, 0.406]; std [0.229, 0.224, 0.225]\\
Data augmentations (train) & RandomHorizontalFlip (p = 0.5); RandomRotation $\pm30^{\circ}$; ColorJitter\\
Optimizer & AdamW (lr = $1\times10^{-4}$)\\
Learning-rate schedule & StepLR (step size 7, $\gamma=0.1$)\\
Loss function & BCEWithLogitsLoss\\
Batch size & 32\\
Number of epochs & 30\\
\bottomrule
\end{tabular}
\end{table}

% Supplementary Table S2 — Clavicle counts
\begin{table}[h]
\centering
\scriptsize
\caption{Counts as a Function of Label, Dataset, and Split for Radiographs of the Clavicle.}
\label{tab:s2_clavicle_counts}
\setlength{\tabcolsep}{4pt}
\renewcommand{\arraystretch}{1.1}
\begin{tabularx}{\textwidth}{@{} X c c c c c c @{}} 
\toprule
\textbf{Label} & \textbf{Incl.\ Train} & \textbf{Excl.\ Train} & \textbf{Incl.\ Val} & \textbf{Excl.\ Val} & \textbf{Internal Test} & \textbf{External Test}\\
\midrule
Fracture (All Locations)                                   & 379 & 379 & 95 & 95 & 121 & 147 \\
Medial Third Fracture                                      & 27  & 22  & 7  & 6  & 7   & 14  \\
Middle Third Fracture                                      & 211 & 207 & 52 & 51 & 66  & 76  \\
Lateral Third Fracture                                     & 148 & 146 & 37 & 36 & 45  & 67  \\
Comminuted or Fragmented Fracture (All Locations)          & 146 & 146 & 36 & 36 & 43  & 59  \\
Displacement                                               & 331 & 329 & 83 & 83 & 98  & 123 \\
Sclerotic Lesion                                           & 6   & 5   & 3  & 3  & 2   & 9   \\
Joint Dislocation (All Locations)                          & 15  & 15  & 4  & 4  & 9   & 16  \\
Joint Subluxation (All Locations)                          & 6   & 6   & 2  & 2  & 6   & 7   \\
Joint Degeneration (All Locations)                         & 48  & 48  & 12 & 12 & 19  & 33  \\
Acromioclavicular Joint -- Joint Space widened             & 39  & 38  & 10 & 10 & 14  & 34  \\
Acromioclavicular Joint -- Subluxation                     & 18  & 18  & 4  & 4  & 6   & 5   \\
Acromioclavicular Joint -- Dislocation                     & 32  & 30  & 7  & 7  & 8   & 16  \\
Acromioclavicular Joint Degeneration                       & 58  & 58  & 15 & 15 & 18  & 28  \\
Swelling or Hematoma                                       & 31  & 30  & 7  & 7  & 12  & 34  \\
Soft Tissue Calcifications                                 & 42  & 41  & 11 & 10 & 5   & 12  \\
Foreign Bodies                                             & 16  & 16  & 5  & 5  & 7   & 13  \\
Ossicles                                                   & 8   & 7   & 2  & 2  & 2   & 4   \\
\bottomrule
\end{tabularx}

\vspace{2pt}
\raggedright \footnotesize \textit{Note:} Incl.\ = inclusive; Excl.\ = exclusive; Train = training set; Val = validation set; Test = test set.
\end{table}

\begin{table}[h]
\caption{Counts as a Function of Label, Dataset, and Split for Radiographs of the Elbow.}
\label{tab:s3_elbow_counts}
\centering
\scriptsize
\setlength{\tabcolsep}{4pt}
\renewcommand{\arraystretch}{1.1}

\begin{tabularx}{\textwidth}{@{} X c c c c c c @{}} 
\toprule
\textbf{Label} & \textbf{Incl.\ Train} & \textbf{Excl.\ Train} & \textbf{Incl.\ Val} & \textbf{Excl.\ Val} & \textbf{Internal Test} & \textbf{External Test}\\
\midrule
Fracture (All Locations)                         & 505 & 379 & 131 & 95 & 162 & 56 \\
Lytic Lesion                                      & 8   & 8   & 3   & 2  & 3   & 2  \\
Sclerotic Lesion                                  & 37  & 34  & 9   & 8  & 12  & 21 \\
Distal Humerus Fracture                           & 71  & 53  & 21  & 13 & 20  & 4  \\
Distal Humerus -- Comminuted or Fragmented        & 21  & 21  & 5   & 3  & 5   & 2  \\
Distal Humerus -- Displacement                    & 29  & 28  & 9   & 8  & 10  & 3  \\
Distal Humerus Fracture -- Extension into Joint   & 13  & 12  & 5   & 3  & 3   & 1  \\
Olecranon Fracture                                & 95  & 82  & 30  & 21 & 25  & 6  \\
Olecranon -- Displaced                            & 45  & 42  & 11  & 10 & 13  & 3  \\
Olecranon -- Comminuted or Fragmented             & 18  & 17  & 6   & 5  & 6   & 3  \\
Coronoid Process Fracture                         & 33  & 22  & 6   & 5  & 6   & 4  \\
Coronoid Process -- Avulsion of the Tip           & 13  & 9   & 4   & 3  & 2   & 2  \\
Ulna Fracture                                     & 59  & 52  & 18  & 13 & 30  & 9  \\
Radial Head Fracture                              & 388 & 312 & 97  & 78 & 102 & 42 \\
Radial Head -- Displaced                          & 67  & 63  & 19  & 17 & 23  & 9  \\
Radial Head -- Comminuted or Fragmented           & 17  & 17  & 5   & 4  & 5   & 5  \\
Radial Neck Fracture                              & 49  & 37  & 13  & 9  & 13  & 4  \\
Radial Neck -- Displaced                          & 4   & 4   & 3   & 2  & 2   & 2  \\
Radius Fracture                                   & 272 & 218 & 65  & 54 & 112 & 44 \\
Joint Subluxation (All Locations)                 & 10  & 8   & 4   & 3  & 2   & 1  \\
Joint Dislocation (All Locations)                 & 33  & 30  & 9   & 8  & 10  & 1  \\
Joint Degeneration (All Locations)                & 215 & 210 & 56  & 53 & 52  & 39 \\
Soft Tissue Calcifications                        & 209 & 207 & 54  & 52 & 61  & 30 \\
Soft Tissue Masses or Mass-like Lesions           & 7   & 6   & 2   & 2  & 2   & 2  \\
Fat Pad Sign                                      & 322 & 299 & 76  & 75 & 95  & 38 \\
Foreign Bodies                                    & 50  & 50  & 12  & 12 & 20  & 8  \\
Ossicles                                          & 75  & 71  & 18  & 18 & 19  & 16 \\
Exostosis                                         & 73  & 72  & 19  & 18 & 57  & 40 \\
\bottomrule
\end{tabularx}

\vspace{2pt}
\raggedright \footnotesize \textit{Note:} Incl.\ = inclusive; Excl.\ = exclusive; Train = training set; Val = validation set; Test = test set.
\end{table}

\begin{table}[h]
\caption{Supplementary Table S4: Counts as a Function of Label, Dataset, and Split for Radiographs of the Thumb.}
\label{tab:s4_thumb_counts}
\centering
\scriptsize
\setlength{\tabcolsep}{4pt}
\renewcommand{\arraystretch}{1.1}

\begin{tabularx}{\textwidth}{@{} X c c c c c c @{}} 
\toprule
\textbf{Label} & \textbf{Incl.\ Train} & \textbf{Excl.\ Train} & \textbf{Incl.\ Val} & \textbf{Excl.\ Val} & \textbf{Internal Test} & \textbf{External Test}\\
\midrule
Fracture (All Locations)                          & 269 & 197 & 60 & 49 & 73 & 55 \\
Comminuted or Fragmented Fracture (All Locations) & 35  & 34  & 10 & 9  & 15 & 13 \\
First Metacarpal Bone Fracture                    & 37  & 22  & 7  & 5  & 8  & 7  \\
Proximal Phalanx Fracture                         & 75  & 50  & 15 & 12 & 16 & 23 \\
Proximal Phalanx -- Comminuted or Fragmented      & 7   & 6   & 5  & 5  & 2  & 3  \\
Distal Phalanx Fracture                           & 163 & 128 & 42 & 34 & 49 & 28 \\
Distal Phalanx -- Comminuted or Fragmented        & 28  & 27  & 8  & 7  & 14 & 10 \\
Joint Subluxation (All Locations)                 & 44  & 38  & 10 & 8  & 18 & 16 \\
Joint Dislocation (All Locations)                 & 19  & 19  & 4  & 4  & 7  & 4  \\
Joint Degeneration (All Locations)                & 163 & 159 & 41 & 41 & 52 & 49 \\
Carpometacarpal Joint -- Subluxation              & 10  & 10  & 4  & 2  & 5  & 3  \\
Carpometacarpal Joint Degeneration                & 101 & 101 & 25 & 25 & 30 & 30 \\
Metacarpophalangeal Joint -- Subluxation          & 24  & 18  & 5  & 5  & 11 & 10 \\
Metacarpophalangeal Joint -- Dislocation          & 6   & 6   & 1  & 1  & 2  & 1  \\
Metacarpophalangeal Joint Degeneration            & 42  & 40  & 10 & 10 & 21 & 20 \\
Interphalangeal Joint -- Subluxation              & 22  & 18  & 4  & 4  & 3  & 2  \\
Interphalangeal Joint -- Dislocation              & 16  & 15  & 3  & 3  & 3  & 4  \\
Interphalangeal Joint Degeneration                & 80  & 78  & 20 & 20 & 19 & 23 \\
Swelling/Dactylitis                               & 81  & 81  & 21 & 20 & 23 & 47 \\
Soft Tissue Calcifications                        & 37  & 34  & 8  & 8  & 7  & 15 \\
Soft Tissues Masses or Mass-like Lesions          & 5   & 5   & 1  & 1  & 1  & 0  \\
Foreign Bodies                                    & 46  & 34  & 9  & 8  & 13 & 7  \\
Ossicles                                          & 106 & 100 & 28 & 25 & 32 & 36 \\
\bottomrule
\end{tabularx}

\vspace{2pt}
\raggedright \footnotesize \textit{Note:} Incl.\ = inclusive; Excl.\ = exclusive; Train = training set; Val = validation set; Test = test set.
\end{table}

\begin{table}[h]

\caption{ Counts of True, False, and Uncertain Labels in the Training and Validation Sets.}
\label{tab:s5_tfu_counts}
\centering
\scriptsize
\setlength{\tabcolsep}{6pt}
\renewcommand{\arraystretch}{1.15}

\begin{tabular}{@{} c c c c @{}} 
\toprule
\textbf{Region} & \textbf{True} & \textbf{False} & \textbf{Uncertain} \\
\midrule
Clavicle & 1,942 & 22,378 & 42 \\
Elbow    & 2,958 & 83,811 & 492 \\
Thumb    & 1,529 & 37,865 & 231 \\
\bottomrule
\end{tabular}
\end{table}

% in preamble, if not already

\begin{table}[h]
\captionsetup{justification=raggedright,singlelinecheck=false}
\caption{ Inclusive vs.\ Exclusive Performance on the Internal Clavicle Test set (Labels with $n\ge10$). 95\% Confidence Intervals in Parentheses.}
\label{tab:s6_clavicle_internal}
\centering
\scriptsize
\setlength{\tabcolsep}{3.5pt}
\renewcommand{\arraystretch}{1.1}

\resizebox{\textwidth}{!}{%
\begin{tabular}{@{} p{0.35\textwidth} r
                *{2}{l}
                *{2}{l}
                *{2}{l}
                *{2}{l} @{}} 
\toprule
\multirow{2}{*}{\textbf{Label}} & 
\multirow{2}{*}{\textbf{N}} &
\multicolumn{2}{c}{\textbf{AUC}} &
\multicolumn{2}{c}{\textbf{Accuracy [\%]}} &
\multicolumn{2}{c}{\textbf{Sensitivity [\%]}} &
\multicolumn{2}{c}{\textbf{Specificity [\%]}}\\
\cmidrule(lr){3-4}\cmidrule(lr){5-6}\cmidrule(lr){7-8}\cmidrule(lr){9-10}
 & & \textbf{Incl} & \textbf{Excl} & \textbf{Incl} & \textbf{Excl} &
   \textbf{Incl} & \textbf{Excl} & \textbf{Incl} & \textbf{Excl} \\
\midrule
Fracture (All Locations) & 121 & 0.95 (0.91, 0.97) & 0.94 (0.90, 0.96) & 87 (82, 91) & 86 (82, 91) & 80 (73, 87) & 91 (85, 96) & 94 (89, 98) & 81 (74, 88) \\
Displacement & 98 & 0.91 (0.87, 0.94) & 0.90 (0.85, 0.94) & 82 (77, 87) & 85 (80, 89) & 84 (76, 90) & 88 (80, 94) & 81 (75, 89) & 83 (77, 89) \\
Middle Third Fracture & 66 & 0.84 (0.78, 0.89) & 0.86 (0.80, 0.91) & 79 (73, 83) & 80 (75, 85) & 73 (62, 82) & 65 (53, 76) & 81 (75, 86) & 86 (81, 91) \\
Lateral Third Fracture & 45 & 0.83 (0.75, 0.89) & 0.83 (0.76, 0.90) & 79 (73, 85) & 61 (55, 67) & 69 (55, 82) & 89 (80, 98) & 81 (75, 87) & 54 (48, 61) \\
Comminuted or Fragmented Fracture (All) & 43 & 0.89 (0.84, 0.93) & 0.89 (0.84, 0.93) & 74 (68, 79) & 76 (71, 82) & 95 (88, 100) & 88 (79, 97) & 69 (62, 76) & 74 (67, 80) \\
Joint Degeneration (All) & 19 & 0.78 (0.65, 0.89) & 0.76 (0.64, 0.87) & 90 (86, 94) & 82 (77, 87) & 26 (8, 48) & 47 (25, 71) & 95 (92, 98) & 85 (79, 89) \\
Acromioclavicular Joint Degeneration & 18 & 0.78 (0.65, 0.89) & 0.77 (0.64, 0.87) & 81 (76, 86) & 76 (71, 82) & 56 (33, 78) & 56 (31, 78) & 83 (77, 88) & 78 (73, 83) \\
Acromioclavicular Joint -- Joint Space Widened & 14 & 0.60 (0.46, 0.72) & 0.63 (0.48, 0.76) & 62 (56, 68) & 70 (64, 76) & 57 (31, 82) & 50 (25, 78) & 62 (56, 68) & 71 (65, 77) \\
Swelling or Hematoma & 12 & 0.59 (0.39, 0.75) & 0.70 (0.55, 0.83) & 45 (39, 51) & 73 (68, 79) & 75 (50, 100) & 42 (13, 71) & 43 (37, 50) & 75 (69, 81) \\
\bottomrule
\end{tabular}%
}
\end{table}

% =========================
% Supplementary Table S7
% =========================
\begin{table}[h]
\caption{ Inclusive vs.\ Exclusive Performance on the External Clavicle Test set (Labels with $n\ge10$). 95\% Confidence Intervals in Parentheses.}
\label{tab:s7_clavicle_external}
\centering
\scriptsize
\setlength{\tabcolsep}{3.5pt}\renewcommand{\arraystretch}{1.1}
\resizebox{\textwidth}{!}{%
\begin{tabular}{@{} p{0.35\textwidth} r *{2}{l} *{2}{l} *{2}{l} *{2}{l} @{}} 
\toprule
\multirow{2}{*}{\textbf{Label}} & \multirow{2}{*}{\textbf{N}} &
\multicolumn{2}{c}{\textbf{AUC}} &
\multicolumn{2}{c}{\textbf{Accuracy [\%]}} &
\multicolumn{2}{c}{\textbf{Sensitivity [\%]}} &
\multicolumn{2}{c}{\textbf{Specificity [\%]}}\\
\cmidrule(lr){3-4}\cmidrule(lr){5-6}\cmidrule(lr){7-8}\cmidrule(lr){9-10}
 & & \textbf{Incl} & \textbf{Excl} & \textbf{Incl} & \textbf{Excl} & \textbf{Incl} & \textbf{Excl} & \textbf{Incl} & \textbf{Excl} \\
\midrule
Fracture (All Locations) & 147 & 0.91 (0.88, 0.94) & 0.91 (0.88, 0.95) & 84 (80, 88) & 86 (82, 90) & 73 (65, 80) & 81 (74, 87) & 95 (92, 98) & 90 (85, 95) \\
Displacement & 123 & 0.93 (0.90, 0.95) & 0.92 (0.88, 0.95) & 84 (80, 89) & 83 (79, 87) & 80 (72, 87) & 70 (62, 77) & 87 (82, 92) & 93 (89, 96) \\
Middle Third Fracture & 76 & 0.90 (0.86, 0.94) & 0.88 (0.84, 0.92) & 85 (81, 89) & 81 (76, 85) & 67 (56, 77) & 50 (39, 61) & 92 (88, 95) & 91 (87, 95) \\
Lateral Third Fracture & 67 & 0.81 (0.75, 0.87) & 0.83 (0.77, 0.88) & 71 (66, 76) & 63 (58, 69) & 73 (63, 84) & 87 (78, 95) & 71 (65, 76) & 57 (51, 63) \\
Comminuted or Fragmented Fracture (All) & 59 & 0.91 (0.87, 0.94) & 0.91 (0.87, 0.94) & 79 (75, 84) & 83 (79, 88) & 92 (83, 98) & 81 (71, 91) & 76 (71, 81) & 84 (79, 89) \\
Acromioclavicular Joint -- Joint Space Widened & 34 & 0.53 (0.42, 0.64) & 0.51 (0.41, 0.62) & 63 (57, 68) & 69 (64, 74) & 44 (27, 61) & 26 (12, 42) & 65 (60, 71) & 74 (69, 80) \\
Swelling or Hematoma & 34 & 0.47 (0.36, 0.58) & 0.54 (0.42, 0.65) & 60 (55, 66) & 82 (77, 86) & 41 (25, 58) & 26 (11, 41) & 63 (57, 69) & 89 (85, 93) \\
Joint Degeneration (All) & 33 & 0.75 (0.68, 0.82) & 0.76 (0.68, 0.83) & 82 (77, 86) & 77 (73, 82) & 27 (13, 42) & 42 (24, 59) & 88 (84, 92) & 82 (77, 86) \\
Acromioclavicular Joint -- Signs of Degeneration & 28 & 0.74 (0.65, 0.82) & 0.74 (0.65, 0.82) & 70 (65, 76) & 74 (69, 79) & 64 (46, 81) & 57 (38, 77) & 71 (65, 77) & 75 (70, 81) \\
\bottomrule
\end{tabular}%
}
\end{table}

% =========================
% Supplementary Table S8
% =========================
\begin{table}[h]

\caption{ Inclusive vs.\ Exclusive Performance on the Internal Elbow Test set (Labels with $n\ge10$). 95\% Confidence Intervals in Parentheses.}
\label{tab:s8_elbow_internal}
\centering
\scriptsize
\setlength{\tabcolsep}{3.5pt}\renewcommand{\arraystretch}{1.1}
\resizebox{\textwidth}{!}{%
\begin{tabular}{@{} p{0.35\textwidth} r *{2}{l} *{2}{l} *{2}{l} *{2}{l} @{}} 
\toprule
\multirow{2}{*}{\textbf{Label}} & \multirow{2}{*}{\textbf{N}} &
\multicolumn{2}{c}{\textbf{AUC}} &
\multicolumn{2}{c}{\textbf{Accuracy [\%]}} &
\multicolumn{2}{c}{\textbf{Sensitivity [\%]}} &
\multicolumn{2}{c}{\textbf{Specificity [\%]}}\\
\cmidrule(lr){3-4}\cmidrule(lr){5-6}\cmidrule(lr){7-8}\cmidrule(lr){9-10}
 & & \textbf{Incl} & \textbf{Excl} & \textbf{Incl} & \textbf{Excl} & \textbf{Incl} & \textbf{Excl} & \textbf{Incl} & \textbf{Excl} \\
\midrule
Fracture (All Locations) & 162 & 0.87 (0.83, 0.90) & 0.88 (0.85, 0.91) & 80 (77, 83) & 79 (76, 82) & 75 (68, 82) & 81 (75, 87) & 81 (78, 84) & 79 (75, 82) \\
Radius Fracture & 112 & 0.86 (0.82, 0.90) & 0.85 (0.82, 0.89) & 79 (76, 82) & 81 (78, 84) & 76 (68, 83) & 74 (66, 81) & 80 (77, 83) & 82 (79, 85) \\
Radial Head Fracture & 102 & 0.86 (0.81, 0.90) & 0.85 (0.81, 0.89) & 84 (81, 86) & 82 (80, 85) & 69 (59, 78) & 69 (60, 77) & 86 (83, 89) & 85 (82, 87) \\
Fat Pad Sign & 95 & 0.78 (0.72, 0.84) & 0.78 (0.72, 0.84) & 78 (75, 81) & 78 (75, 81) & 62 (52, 72) & 63 (54, 73) & 80 (77, 83) & 80 (77, 83) \\
Soft Tissue Calcifications & 61 & 0.78 (0.71, 0.83) & 0.76 (0.68, 0.83) & 72 (69, 75) & 67 (63, 70) & 67 (55, 79) & 75 (64, 86) & 73 (69, 76) & 66 (62, 69) \\
Exostosis & 57 & 0.82 (0.76, 0.87) & 0.81 (0.76, 0.86) & 67 (63, 70) & 61 (58, 65) & 77 (66, 88) & 84 (74, 93) & 66 (62, 69) & 60 (56, 63) \\
Joint Degeneration (All) & 52 & 0.81 (0.74, 0.86) & 0.82 (0.75, 0.87) & 79 (76, 82) & 71 (68, 74) & 69 (56, 81) & 77 (65, 88) & 80 (76, 82) & 70 (67, 74) \\
Ossicles & 19 & 0.62 (0.48, 0.74) & 0.61 (0.48, 0.72) & 54 (51, 58) & 91 (89, 93) & 63 (40, 83) & 11 (0, 27) & 54 (51, 58) & 93 (91, 95) \\
Sclerotic Lesion & 12 & 0.80 (0.66, 0.92) & 0.85 (0.74, 0.94) & 83 (80, 86) & 70 (67, 73) & 67 (38, 92) & 83 (56, 100) & 83 (80, 86) & 70 (66, 73) \\
\bottomrule
\end{tabular}%
}
\end{table}

% =========================
% Supplementary Table S9
% =========================
\begin{table}[h]
\caption{ Inclusive vs.\ Exclusive Performance on the External Elbow Test set (Labels with $n\ge10$). 95\% Confidence Intervals in Parentheses.}
\label{tab:s9_elbow_external}
\centering
\scriptsize
\setlength{\tabcolsep}{3.5pt}\renewcommand{\arraystretch}{1.1}
\resizebox{\textwidth}{!}{%
\begin{tabular}{@{} p{0.35\textwidth} r *{2}{l} *{2}{l} *{2}{l} *{2}{l} @{}} 
\toprule
\multirow{2}{*}{\textbf{Label}} & \multirow{2}{*}{\textbf{N}} &
\multicolumn{2}{c}{\textbf{AUC}} &
\multicolumn{2}{c}{\textbf{Accuracy [\%]}} &
\multicolumn{2}{c}{\textbf{Sensitivity [\%]}} &
\multicolumn{2}{c}{\textbf{Specificity [\%]}}\\
\cmidrule(lr){3-4}\cmidrule(lr){5-6}\cmidrule(lr){7-8}\cmidrule(lr){9-10}
 & & \textbf{Incl} & \textbf{Excl} & \textbf{Incl} & \textbf{Excl} & \textbf{Incl} & \textbf{Excl} & \textbf{Incl} & \textbf{Excl} \\
\midrule
Fracture (All Locations) & 56 & 0.85 (0.79, 0.90) & 0.85 (0.79, 0.90) & 76 (71, 81) & 76 (71, 81) & 70 (57, 81) & 86 (75, 94) & 77 (72, 83) & 74 (68, 79) \\
Radius Fracture & 44 & 0.84 (0.76, 0.91) & 0.88 (0.81, 0.94) & 75 (70, 80) & 80 (76, 84) & 75 (63, 87) & 82 (70, 93) & 75 (70, 81) & 80 (75, 84) \\
Radial Head Fracture & 42 & 0.87 (0.80, 0.93) & 0.89 (0.84, 0.94) & 84 (80, 88) & 85 (81, 89) & 69 (55, 83) & 83 (71, 93) & 87 (83, 91) & 85 (81, 89) \\
Exostosis & 40 & 0.86 (0.80, 0.91) & 0.85 (0.78, 0.91) & 73 (68, 78) & 58 (52, 63) & 95 (88, 100) & 93 (84, 100) & 70 (65, 75) & 52 (46, 58) \\
Fat Pad Sign & 38 & 0.78 (0.68, 0.86) & 0.78 (0.69, 0.86) & 68 (62, 73) & 69 (64, 74) & 71 (56, 85) & 71 (55, 85) & 67 (61, 73) & 69 (63, 74) \\
Joint Degeneration (All) & 39 & 0.84 (0.77, 0.92) & 0.84 (0.77, 0.91) & 81 (76, 85) & 71 (66, 77) & 74 (60, 88) & 82 (69, 93) & 82 (77, 87) & 70 (64, 75) \\
Soft Tissue Calcifications & 30 & 0.61 (0.49, 0.71) & 0.63 (0.52, 0.73) & 75 (70, 80) & 68 (63, 74) & 37 (19, 53) & 53 (36, 71) & 80 (74, 84) & 70 (65, 75) \\
Sclerotic Lesion & 21 & 0.65 (0.53, 0.77) & 0.63 (0.52, 0.75) & 72 (66, 77) & 60 (55, 65) & 43 (19, 67) & 52 (32, 75) & 74 (69, 79) & 61 (55, 66) \\
Ossicles & 16 & 0.78 (0.66, 0.90) & 0.75 (0.60, 0.87) & 60 (55, 66) & 92 (89, 95) & 75 (53, 94) & 19 (0, 38) & 60 (54, 65) & 96 (94, 99) \\
\bottomrule
\end{tabular}%
}
\end{table}

% =========================
% Supplementary Table S10
% =========================
\begin{table}[h]
\caption{ Inclusive vs.\ Exclusive Performance on the Internal Thumb Test set (Labels with $n\ge10$). 95\% Confidence Intervals in Parentheses.}
\label{tab:s10_thumb_internal}
\centering
\scriptsize
\setlength{\tabcolsep}{3.5pt}\renewcommand{\arraystretch}{1.1}
\resizebox{\textwidth}{!}{%
\begin{tabular}{@{} p{0.35\textwidth} r *{2}{l} *{2}{l} *{2}{l} *{2}{l} @{}} 
\toprule
\multirow{2}{*}{\textbf{Label}} & \multirow{2}{*}{\textbf{N}} &
\multicolumn{2}{c}{\textbf{AUC}} &
\multicolumn{2}{c}{\textbf{Accuracy [\%]}} &
\multicolumn{2}{c}{\textbf{Sensitivity [\%]}} &
\multicolumn{2}{c}{\textbf{Specificity [\%]}}\\
\cmidrule(lr){3-4}\cmidrule(lr){5-6}\cmidrule(lr){7-8}\cmidrule(lr){9-10}
 & & \textbf{Incl} & \textbf{Excl} & \textbf{Incl} & \textbf{Excl} & \textbf{Incl} & \textbf{Excl} & \textbf{Incl} & \textbf{Excl} \\
\midrule
Fracture (All Locations) & 73 & 0.70 (0.63, 0.77) & 0.71 (0.64, 0.78) & 77 (73, 81) & 63 (58, 68) & 51 (39, 62) & 66 (55, 77) & 83 (79, 87) & 63 (57, 68) \\
Joint Degeneration (All) & 52 & 0.89 (0.84, 0.93) & 0.88 (0.81, 0.93) & 83 (80, 87) & 80 (76, 84) & 85 (74, 93) & 83 (71, 92) & 83 (79, 87) & 79 (75, 84) \\
Distal Phalanx Fracture & 49 & 0.75 (0.66, 0.84) & 0.77 (0.69, 0.85) & 80 (76, 84) & 77 (73, 81) & 49 (34, 62) & 65 (52, 78) & 84 (80, 88) & 78 (74, 83) \\
Ossicles & 32 & 0.64 (0.53, 0.73) & 0.66 (0.56, 0.76) & 56 (51, 60) & 60 (55, 65) & 59 (41, 76) & 63 (44, 80) & 55 (50, 60) & 60 (55, 65) \\
Carpometacarpal Joint Degeneration & 30 & 0.91 (0.85, 0.96) & 0.90 (0.85, 0.95) & 81 (77, 85) & 81 (77, 85) & 87 (74, 97) & 87 (74, 97) & 81 (77, 85) & 81 (77, 85) \\
Swelling/Dactylitis & 23 & 0.68 (0.55, 0.81) & 0.73 (0.63, 0.81) & 89 (86, 92) & 75 (71, 79) & 26 (10, 44) & 52 (30, 73) & 93 (91, 96) & 76 (72, 80) \\
Metacarpophalangeal Joint Degeneration & 21 & 0.88 (0.78, 0.95) & 0.89 (0.79, 0.96) & 80 (77, 84) & 92 (89, 94) & 81 (63, 96) & 67 (44, 86) & 80 (76, 84) & 93 (90, 95) \\
Interphalangeal Joint Degeneration & 19 & 0.85 (0.76, 0.93) & 0.84 (0.73, 0.92) & 75 (71, 79) & 79 (75, 83) & 84 (65, 100) & 74 (53, 93) & 75 (71, 79) & 79 (75, 83) \\
Joint Subluxation & 18 & 0.72 (0.57, 0.85) & 0.69 (0.56, 0.82) & 92 (89, 94) & 92 (89, 94) & 11 (0, 28) & 11 (0, 27) & 96 (93, 98) & 96 (94, 98) \\
Proximal Phalanx Fracture & 16 & 0.61 (0.47, 0.74) & 0.65 (0.53, 0.77) & 35 (31, 40) & 91 (88, 93) & 81 (58, 100) & 13 (0, 31) & 33 (28, 38) & 94 (92, 96) \\
Comminuted or Fragmented Fracture (All) & 15 & 0.80 (0.63, 0.94) & 0.86 (0.71, 0.98) & 87 (83, 90) & 76 (72, 80) & 60 (36, 85) & 80 (56, 100) & 88 (85, 91) & 76 (72, 80) \\
Distal Phalanx -- Comminuted or Fragmented Fracture & 14 & 0.85 (0.69, 0.97) & 0.89 (0.74, 0.99) & 88 (85, 92) & 91 (88, 94) & 71 (45, 100) & 86 (67, 100) & 89 (86, 92) & 91 (88, 94) \\
Metacarpophalangeal Joint Subluxation & 11 & 0.59 (0.41, 0.75) & 0.61 (0.39, 0.82) & 65 (60, 69) & 87 (84, 90) & 55 (25, 86) & 27 (0, 56) & 65 (60, 69) & 89 (86, 92) \\
\bottomrule
\end{tabular}%
}
\end{table}
% =========================
% Supplementary Table S11
% =========================
\begin{table}[h]

\caption{ Inclusive vs.\ Exclusive Performance on the External Thumb Test set (Labels with $n\ge10$). 95\% Confidence Intervals in Parentheses.}
\label{tab:s11_thumb_external}
\centering
\scriptsize
\setlength{\tabcolsep}{3.5pt}\renewcommand{\arraystretch}{1.1}
\resizebox{\textwidth}{!}{%
\begin{tabular}{@{} p{0.35\textwidth} r *{2}{l} *{2}{l} *{2}{l} *{2}{l} @{}} 
\toprule
\multirow{2}{*}{\textbf{Label}} & \multirow{2}{*}{\textbf{N}} &
\multicolumn{2}{c}{\textbf{AUC}} &
\multicolumn{2}{c}{\textbf{Accuracy [\%]}} &
\multicolumn{2}{c}{\textbf{Sensitivity [\%]}} &
\multicolumn{2}{c}{\textbf{Specificity [\%]}}\\
\cmidrule(lr){3-4}\cmidrule(lr){5-6}\cmidrule(lr){7-8}\cmidrule(lr){9-10}
 & & \textbf{Incl} & \textbf{Excl} & \textbf{Incl} & \textbf{Excl} & \textbf{Incl} & \textbf{Excl} & \textbf{Incl} & \textbf{Excl} \\
\midrule
Fracture (All Locations) & 55 & 0.62 (0.54, 0.70) & 0.65 (0.57, 0.73) & 75 (69, 79) & 69 (64, 74) & 27 (16, 38) & 45 (33, 58) & 85 (81, 89) & 75 (69, 80) \\
Joint Degeneration (All) & 49 & 0.90 (0.85, 0.95) & 0.91 (0.86, 0.95) & 83 (79, 87) & 71 (65, 76) & 90 (81, 98) & 92 (83, 100) & 82 (77, 87) & 67 (60, 73) \\
Swelling/Dactylitis & 47 & 0.57 (0.50, 0.66) & 0.60 (0.51, 0.69) & 78 (73, 82) & 62 (57, 68) & 6 (0, 15) & 53 (39, 68) & 91 (87, 94) & 64 (58, 70) \\
Ossicles & 36 & 0.42 (0.33, 0.53) & 0.46 (0.36, 0.56) & 31 (26, 36) & 38 (33, 43) & 58 (41, 73) & 56 (39, 73) & 27 (22, 32) & 36 (30, 41) \\
Carpometacarpal Joint Degeneration & 30 & 0.90 (0.84, 0.95) & 0.90 (0.85, 0.95) & 81 (77, 85) & 76 (71, 80) & 77 (61, 91) & 93 (83, 100) & 81 (77, 86) & 74 (68, 79) \\
Distal Phalanx Fracture & 28 & 0.67 (0.58, 0.76) & 0.68 (0.58, 0.79) & 84 (80, 88) & 80 (75, 84) & 14 (3, 29) & 39 (23, 60) & 91 (87, 94) & 84 (79, 88) \\
Interphalangeal Joint Degeneration & 23 & 0.89 (0.82, 0.94) & 0.89 (0.83, 0.94) & 77 (73, 82) & 77 (73, 81) & 91 (78, 100) & 91 (78, 100) & 76 (71, 81) & 76 (71, 81) \\
Proximal Phalanx Fracture & 23 & 0.46 (0.33, 0.58) & 0.53 (0.41, 0.65) & 36 (31, 42) & 91 (87, 94) & 65 (45, 84) & 0 (0, 0) & 34 (28, 40) & 98 (96, 100) \\
Metacarpophalangeal Joint Degeneration & 20 & 0.85 (0.75, 0.92) & 0.85 (0.77, 0.93) & 84 (79, 88) & 89 (86, 92) & 80 (62, 95) & 40 (17, 63) & 84 (80, 88) & 93 (89, 95) \\
Joint Subluxation & 16 & 0.71 (0.53, 0.86) & 0.70 (0.53, 0.88) & 94 (91, 96) & 93 (90, 96) & 13 (0, 33) & 13 (0, 33) & 98 (97, 100) & 98 (96, 99) \\
Comminuted or Fragmented Fracture (All) & 13 & 0.69 (0.56, 0.81) & 0.73 (0.64, 0.81) & 85 (81, 89) & 73 (67, 77) & 15 (0, 37) & 46 (18, 75) & 89 (84, 92) & 74 (69, 78) \\
Distal Phalanx -- Comminuted or Fragmented Fracture & 10 & 0.66 (0.47, 0.82) & 0.72 (0.58, 0.84) & 86 (82, 90) & 89 (85, 92) & 10 (0, 33) & 10 (0, 36) & 88 (84, 92) & 92 (89, 95) \\
Metacarpophalangeal Joint Subluxation & 10 & 0.64 (0.44, 0.83) & 0.61 (0.36, 0.85) & 71 (66, 76) & 93 (90, 96) & 50 (20, 83) & 30 (0, 64) & 72 (67, 77) & 95 (92, 97) \\
\bottomrule
\end{tabular}%
}
\end{table}

\begin{figure}[h]
\centering
\includegraphics[width=1\linewidth]{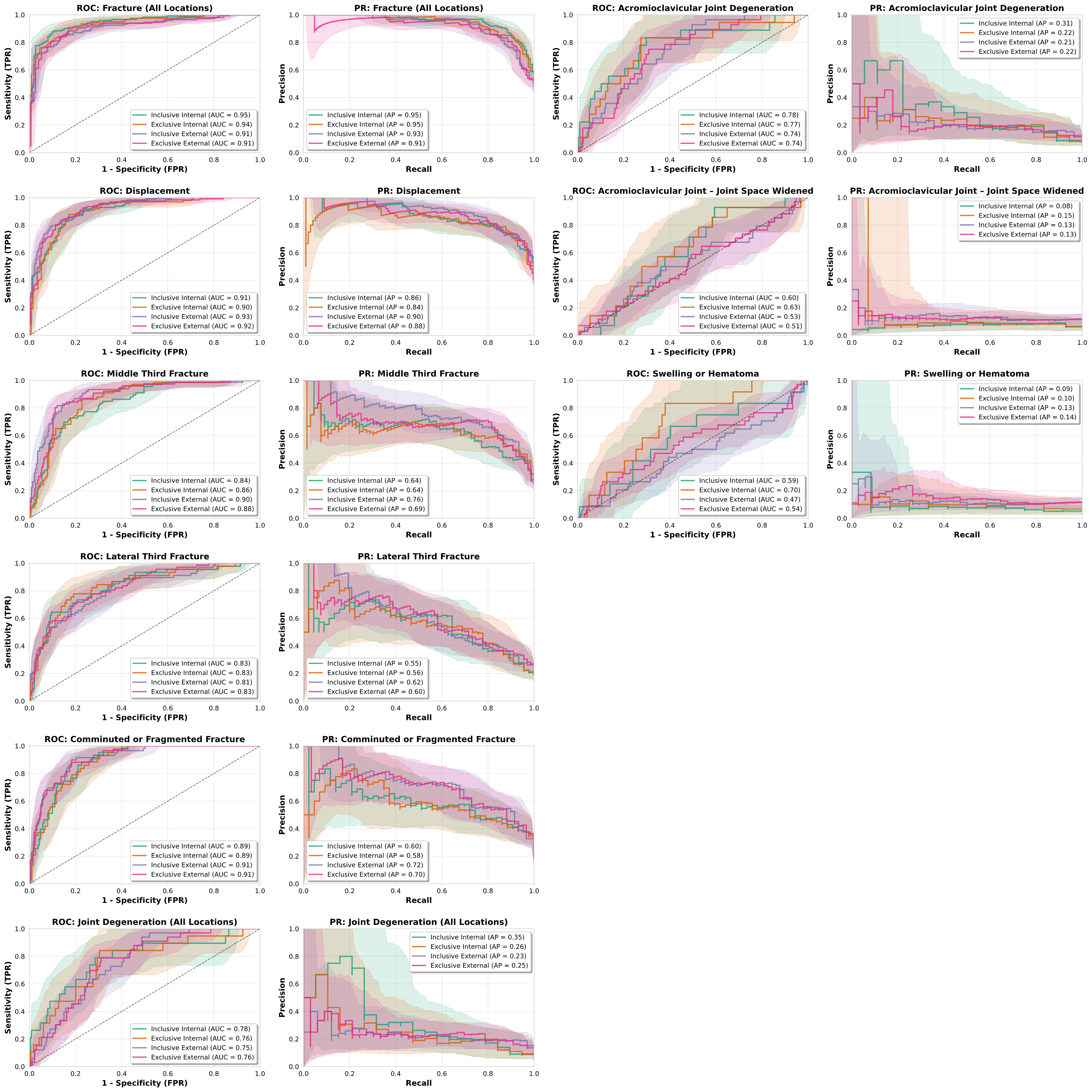}
\caption{Receiver-operating characteristic -and precision recall-curves of the Inclusive and Exclusive Models for the Clavicle for all Labels with n$\geq$10 in Both Datasets.}
\label{fig:s1_clavicle_roc}
\end{figure}

\begin{figure}[h]
\centering
\includegraphics[width=1\linewidth]{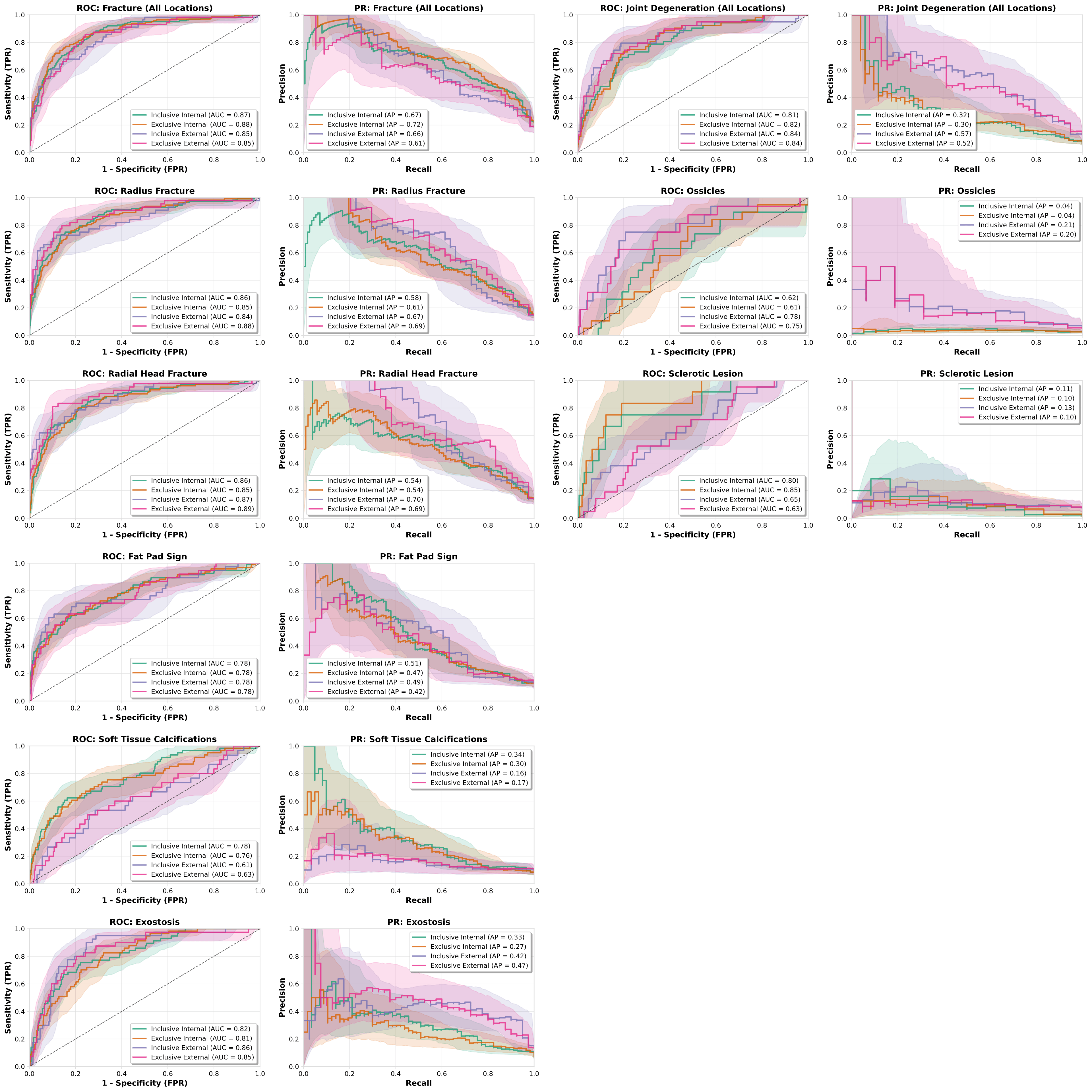}
\caption{Receiver-operating characteristic -and precision recall-curves of the Inclusive and Exclusive Models for the Elbow for all Labels with n$\geq$10 in Both Datasets.}
\label{fig:s1_elbow_roc}
\end{figure}

\begin{figure}[h]
\centering
\includegraphics[width=1\linewidth]{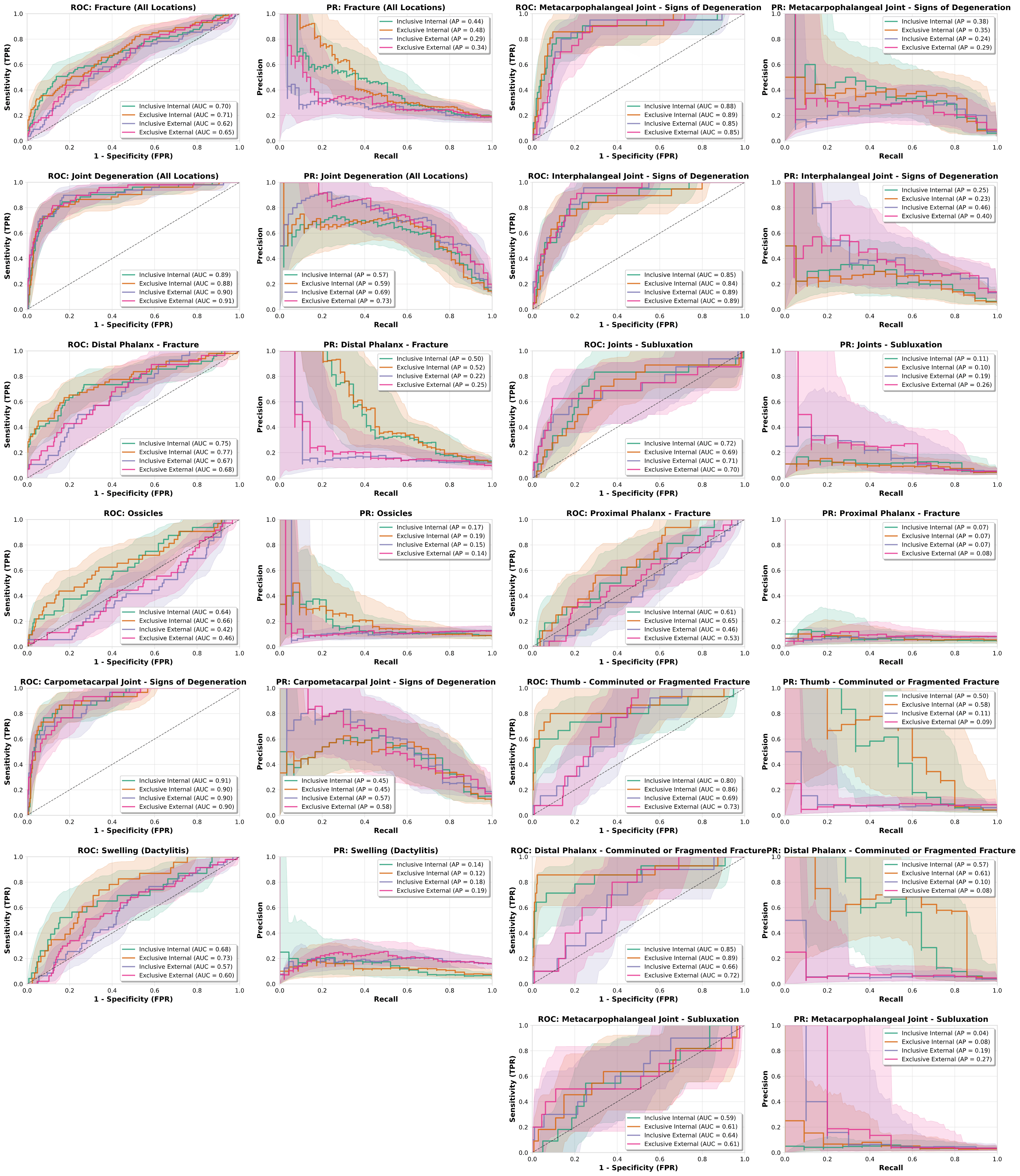}
\caption{Receiver-operating characteristic -and precision recall-curves of the Inclusive and Exclusive Models for the Thumb for all Labels with n$\geq$10 in Both Datasets.}
\label{fig:s1_thumb_roc}
\end{figure}
\end{document}